\documentclass[12pt]{article}

\usepackage{scicite}

\usepackage{times}
\usepackage{xr-hyper}
\usepackage{hyperref} 

\makeatletter

\newenvironment{sciabstract}{%
\begin{quote} \bf}
{\end{quote}}

\title{Accelerating Antimicrobial Discovery with Controllable Deep Generative Models and Molecular Dynamics}

\author
{Payel Das,$^{1,2\ast}$ Tom Sercu,$^{1}$ Kahini Wadhawan,$^{1,\dagger}$ Inkit Padhi,$^{1,\dagger}$ \\
 Sebastian Gehrmann,$^{3,\dagger}$ Flaviu Cipcigan,$^{4}$ Vijil Chenthamarakshan,$^{1}$ 
 \\Hendrik Strobelt,$^{5}$ Cicero dos Santos,$^{1}$ Pin-Yu Chen,$^{1}$ \\
Yi Yan Yang,$^{6}$ Jeremy P. K. Tan,$^{6}$ James Hedrick,$^{7}$ Jason Crain,$^{4,8}$  \\Aleksandra Mojsilovic$^{1}$
\\
\normalsize{$^{1}$IBM Thomas J Watson Research Center,  Yorktown Heights, NY 10598, USA}\\
\normalsize{$^{2}$Department of Applied Physics and Mathematics, Columbia University,}\\
\normalsize{New York, NY 10027, USA}\\
\normalsize{$^{3}$Harvard John A. Paulson School of Engineering and Applied Sciences, Cambridge,}\\
\normalsize{MA 02138. USA}\\
\normalsize{$^{4}$IBM Research Europe, The Hartree Centre STFC Laboratory, Warrington WA4 4AD, UK}\\
\normalsize{$^{5}$IBM Research, MIT-IBM Watson AI Lab,  Cambridge, MA 02142, USA}\\
\normalsize{$^{6}$Institute of Bioengineering and Nanotechnology, Singapore 138669, Singapore}\\
\normalsize{$^{7}$IBM Research, Almaden Research Center,
San Jose, CA 95120, USA}\\
\normalsize{$^{8}$Department of Biochemistry, University of Oxford, Oxford, OX1 3QU, UK}\\
\\
\normalsize{$^\ast$Correspondence to: daspa@us.ibm.com.}
\\
\normalsize{$^\dagger$ These authors contributed equally.}
\\
\normalsize{T. Sercu is currently at Facebook AI Research.} \\
\normalsize{S. Gehrmann was an intern at IBM Research during this work.} \\
\normalsize{C. dos Santos is currently at Amazon Web Services.}
}

\usepackage{amsmath,amsfonts,bm}

\def\eqref#1{equation~\ref{#1}}

\def\1{\bm{1}}

\def\rva{{\mathbf{a}}}

\def\rvx{{\mathbf{x}}}

\def\rvz{{\mathbf{z}}}

\DeclareMathAlphabet{\mathsfit}{\encodingdefault}{\sfdefault}{m}{sl}
\SetMathAlphabet{\mathsfit}{bold}{\encodingdefault}{\sfdefault}{bx}{n}

\def\gL{{\mathcal{L}}}

\newcommand{\E}{\mathbb{E}}

\newcommand{\normltwo}{L^2}

\usepackage{graphicx}
\usepackage{caption}
\usepackage{subcaption}
\usepackage{multirow}
\usepackage{booktabs}
\usepackage{enumerate}
\graphicspath{{figures/}}
\usepackage[x11names]{xcolor}
\usepackage[left]{lineno}

\definecolor{cb_orange}{rgb}{1.0,0.51,0.0}
\definecolor{cb_blue}{rgb}{0.22,0.49,0.72}
\definecolor{cb_green}{rgb}{0.3,0.67,0.29}
\definecolor{cb_red}{rgb}{0.89,0.1,0.11}
\definecolor{cb_purple}{rgb}{0.6, 0.31, 0.64}
\definecolor{cb_brown}{rgb}{0.6, 0.4, 0.2}
\definecolor{cb_crimson}{rgb}{0.86, 0.08, 0.24}
\definecolor{cb_bblue}{rgb}{1.0, 0.44, 0.37}

\definecolor{our_blue}{RGB}{68,150,255}
\definecolor{our_gray}{RGB}{127,127,127}
\definecolor{our_orange}{RGB}{255,134,53}

\date{}

\begin{document} 


\baselineskip24pt

\maketitle 
\section*{Summary}
Synergistic use of controlled sampling from a generative autoencoder and molecular simulation-based screening accelerates design of novel, safe, and potent antimicrobial peptides.


\begin{sciabstract}

\section*{Abstract}

De novo therapeutic design is challenged by a vast chemical repertoire and multiple constraints, e.g., high broad-spectrum potency and low toxicity. We propose CLaSS (Controlled Latent attribute Space Sampling) — an efficient computational method for attribute-controlled generation of molecules, which leverages guidance from classifiers trained on an informative latent space of molecules modeled using a deep generative autoencoder.  We screen the generated molecules for additional key attributes by using deep learning classifiers in conjunction with novel features derived from atomistic simulations. The proposed approach is demonstrated for designing non-toxic antimicrobial peptides (AMPs) with strong broad-spectrum potency, which are emerging drug candidates for tackling antibiotic resistance. Synthesis and testing of only twenty designed sequences identified two novel and minimalist AMPs with high potency against diverse Gram-positive and Gram-negative pathogens, including one multidrug-resistant and one antibiotic-resistant K. pneumoniae, via membrane pore formation.  Both antimicrobials exhibit low in vitro and in vivo toxicity and mitigate the onset of drug resistance. The proposed approach thus presents a viable path for faster and efficient discovery of potent and selective broad-spectrum antimicrobials.

\end{sciabstract}

\section*{Introduction}

\textit{De novo} therapeutic molecule design remains a cost and time-intensive process: It typically requires more than ten years and \$2-3 B USD for a new drug to reach the market, and the success  rate  could be as low as $<1$\%. \cite{dimasi2016innovation, desselle2017institutional}. Efficient computational strategies for targeted generation and screening of molecules with desired therapeutic properties are therefore urgently required. As a specific example, we consider here the antimicrobial peptide (AMP) design problem.
Antimicrobial peptides are emerging drug candidates for tackling antibiotic resistance, 
one of the biggest threats in global health, food security, and development. Patients at a higher risk from drug-resistant pathogens are also more vulnerable to illness from viral lung infections like influenza, severe acute respiratory syndrome (SARS), and COVID-19. 
Drug-resistant diseases claim 700,000 lives a year globally~\cite{UN_2019}, which is expected to rise to 10 million deaths per year by 2050 based on current trends~\cite{oneill_2016}. Of particular concern are multidrug-resistant Gram-negative bacteria~\cite{WHO_2019}. 
Antimicrobial peptides that are believed to be the antibiotic of last resort 
are typically 12-50 amino acids long and produced by multiple higher-order organisms to combat invading microorganisms. 
Due to their exceptional structural and functional variety  \cite{Powers2003}, promising activity, and low tendency to induce (or even reduce) resistance, 
natural AMPs have been proposed as promising alternatives to traditional antibiotics and as potential next-generation antimicrobial agents \cite{Mahlapuu2016}. 
Most reported antimicrobials are cationic  and amphiphilic in nature, and possess properties thought to be crucial for insertion into and disruption of bacterial membrane 
~\cite{Mahlapuu2016}.

Rational methods for novel therapeutic peptide design,
both in wet lab and \textit{in silico},  
heavily rely upon  
structure-activity relationship (SAR) studies \cite{chen2019simulation, torres2018structure, tucker2018discovery, field2013saturation, fjell2012designing, li2017membrane}. Such methods struggle with 
the prohibitively large molecular space, 
complex structure-function relationships, and multiple competing constraints  
such as activity, toxicity, synthesis cost, and stability associated with the design task.
Recently, artificial intelligence (AI) methods, in particular, statistical learning and optimization-based approaches, have shown promise in designing small- and macro-molecules, including antimicrobial peptides. 
A  comprehensive review of computational methodologies for AMP design can be found in ref \cite{cardoso2020computer}.  
A conventional approach is to build a predictive model that estimates the properties of a given molecule, which is then used for candidate screening \cite{jenssen2008qsar, vishnepolsky2019novo, maccari2013antimicrobial, meher2017predicting, thomas2010camp, witten2019deep,xiao2013iamp,veltri2018deep}. Either a manually selected or automatically learned set of compositional, structural, or physicochemical features, or direct sequence is used to build the predictive model. A candidate is typically obtained by combinatorial enumeration of chemically plausible fragments (or sub-sequences) followed by random selection,  from an existing molecular library, or modification thereof. Sequence optimization using genetic algorithms \cite{porto2018silico, fjell2011optimization}, pattern insertion \cite{porto2018joker}, or sub-graph matching \cite{nagarajan2019omega76} has been also used in the development of new drugs, which requires  selection of initial (or template) sequence  and/or defining patterns.  An alternative  is to develop a generative model 
for automated \textit{de novo} design of novel molecules with user-specified properties.  Deep learning-based architectures, such as neural language models as well as deep generative neural networks, have emerged as a popular choice  
\cite{muller2018recurrent, grisoni2018designing, gupta2018generative,
gomez1610automatic, jin2018junction, blaschke2018application, chan2019advancing, sanchez2018inverse, nagarajan2018computational}.  Probabilistic autoencoders \cite{hinton2006reducing, kingma2013auto}, a powerful class of deep generative models,  
have been used for 
this design task, which 
learn a bidirectional mapping of the input molecules (and their attributes) to a continuous latent space.

 Earlier deep generative models for targeted generation have often limited the learning to a fixed library of molecules with desired attributes, to restrict the exhaustive search to a defined section of the chemical space. Such an approach can affect the novelty as well as the validity of the generated molecules, as the fixed library represents a small portion of the combinatorial molecular space \cite{porto2018silico}.
Alternative methods include Bayesian optimization (BO) on a learned latent space \cite{gomez1610automatic},  reinforcement learning (RL) \cite{guimaraes2017objective, popova2018deep}, or semi-supervised learning (SS) \cite{kang2018conditional}. 
However, those approaches require surrogate model fitting (as in  BO), optimal policy learning (as in RL),  or minimizing attribute-specific loss objectives (as in SS),  which suffers from additional computational complexity.  As a result, controlling attribute(s) of designed molecules efficiently continues to remain a non-trivial task. 

To tackle these challenges, we propose a computational framework for targeted design and screening of molecules, which combines  attribute-controlled deep generative models and physics-driven simulations. 
For targeted generation, we propose Conditional Latent (attribute) Space Sampling -- CLaSS 
that leverages guidance from attribute classifier(s) trained on the latent space of the system of interest and uses a rejection sampling scheme for generating molecules with desired attributes. 
CLaSS has several advantages, as it is  efficient and easily repurposable in comparison to existing machine learning algorithms  for targeted generation. 
To encourage novelty and validity of designed sequences, we performed CLaSS on the latent space of a  deep generative autoencoder that was trained on a larger dataset consisting of all known peptide sequences, instead of a limited number of known antimicrobials. Extensive analyses showed that the resulting latent space is informative of peptide properties. As a result, the antimicrobial peptides generated from this informative space are novel, diverse, valid, and optimized. 

Although several antimicrobial peptides are in clinical trials \cite{Mahlapuu2016}, the future design of novel AMP therapeutics requires minimizing the high production cost due to longer sequence length, proteolytic degradation, poor solubility, and off-target toxicity.  A rational path for resolving these problems is to design  short peptides as a minimal physical model 
\cite{Losasso2019, Cipcigan2018} that captures the high selectivity of natural AMPs. That is, maximizing antimicrobial activity, while minimizing toxicity towards the host.

\begin{figure}
\centering
\includegraphics[width=\textwidth]{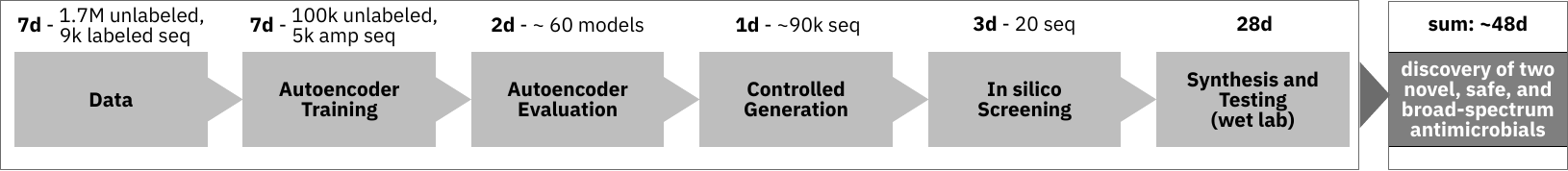}
\caption{Overview and timeline of the proposed AI-driven approach for accelerated antimicrobial design.     
}
\label{fig:Overview}
\end{figure}

To account for these additional key requirements, such as broad-spectrum potency and low toxicity,  we further provide an efficient \textit{in silico} screening method that uses deep learning classifiers augmented with high-throughput physics-driven molecular simulations (Fig. \ref{fig:Overview}). To our knowledge, this is the first computational approach for \textit{de novo} antimicrobial design that explicitly accounts for broad-spectrum potency and low toxicity, and performs experimental verification of those properties.   Synthesis of  20 candidate sequences (from a pool of $\sim$ 90,000 generated sequences) that passed the screening enabled discovery of two novel and short length peptides with experimentally validated strong antimicrobial activity against diverse pathogens,  including a hard-to-treat multidrug-resistant Gram-negative \textit{K. pneumoniae}.  Importantly, both sequences demonstrated  low \textit{in vitro} hemolysis (HC50) and \textit{in vivo} lethal (LD50) toxicity.  Circular dichroism experiments  revealed the amphiphilic helical topology of the two novel cationic AMPs.  All-atom simulations of helical YI12 and FK13 show distinct modes of early lipid membrane interaction. Wet lab experiments further confirmed their bactericidal nature; while live imaging with confocal microscopy showed bacterial membrane permeability. Both peptides displayed low propensity to induce resistance onset in \textit{E. coli} compared to imipenem, an existing antibiotic. No cross-resistance was seen for either YI12 or FK13, when tested using a polymyxin-resistant strain. Taken together, both YI12 and FK13 appear promising therapeutic candidates that deserve further investigation.
The present strategy, therefore, provides an efficient \textit{de novo} approach for discovering novel, broad-spectrum, and low-toxic antimicrobials with therapeutic potential at 10\% success rate and rapid (48 days) pace. 

\section*{Results}




\subsection*
{Peptide autoencoder}
\label{sec:lvm}

For modeling the peptide latent space, we used generative models based on a deep autoencoder  \cite{hinton2006reducing, kingma2013auto} 
composed of two neural networks, an encoder and a decoder.   The encoder $q_{\phi}(\rvz|\rvx)$ parameterized with $\phi$ learns to map the input $\rvx$ to a variational distribution, and the decoder $p_{\theta}(\rvx|\rvz)$ parameterized with $\theta$ aims to reconstruct the input $\rvx$ given the latent vector $\rvz$ from the learned distribution, as illustrated in Fig. \ref{fig:overview-gen}A. Variational Autoencoder (VAE), the most popular model in this family \cite{kingma2013auto},  assumes \emph{latent variable} $\rvz \sim p(\rvz)$ and follows a simple prior (\textit{e.g.} Gaussian) distribution.  And the decoder then produces a distribution over sequences given the continuous representation $\rvz$. 
Thus, the generative process is specified as: $p(\rvx) = \int p(\rvz) p_\theta(\rvx | \rvz) d\rvz$ where we integrate out the latent variable. 
An alternative (and supposedly improved variant)  of standard VAE is 
Wasserstein Autoencoder (WAE) (For details, see SI). 
Within the VAE/WAE framework, the peptide generation is formulated as a density modeling problem, \textit{i.e.} estimating $p(\rvx)$ where $\rvx$ are short variable-length strings of amino acids. The density estimation procedure has to assign a high likelihood to known peptides. Therefore, the model generalization implies that plausible novel peptides can be generated from regions with a high probability density under the model.  Peptide sequences are presented as text strings composed of 20 natural amino acid characters. Only  sequences with length $\le  25$ were considered for model training and generation, as short AMPs are desired.

However, instead of learning a model only over known AMP sequences, one can learn a model over all  peptide sequences reported in the UniProt database \cite{uniprot} -- an extensive database of protein/peptide sequences that may or may not have an annotation. For example, the number of annotated AMP sequences is $\sim$ 9000, and peptide sequences in Uniprot is  $\sim$ 1.7M, when a sequence length up to 50 is considered.  
Therefore, we learn a density model of over all known peptides sequences, in this work. The fact also inspires this approach, that unsupervised representation learning by pre-training on a large corpus has recently led to impressive results for downstream tasks in text and speech ~\cite{ElmoPeters:2018, gpt2:radford2019language, cove:mccann2017learned, devlin2018bert}, as well as in protein biology \cite{rao2019evaluating,  Madani2020.03.07.982272}. 
Additionally, in contrast to similar models for protein sequence generation \cite{riesselman2017deep}, we do not restrict ourselves to learning the density associated with a single protein family or a specific 3D fold.
Instead, we learn a global model, over all known short peptide sequences expressed in different organisms. 
This global approach should enable meaningful density modeling across multiple families, the interpolation between them, better learning of the ``grammar'' of plausible peptides, and exploration beyond known antimicrobial templates, as shown next.

\begin{figure}
     \centering
     \includegraphics[width=\textwidth]{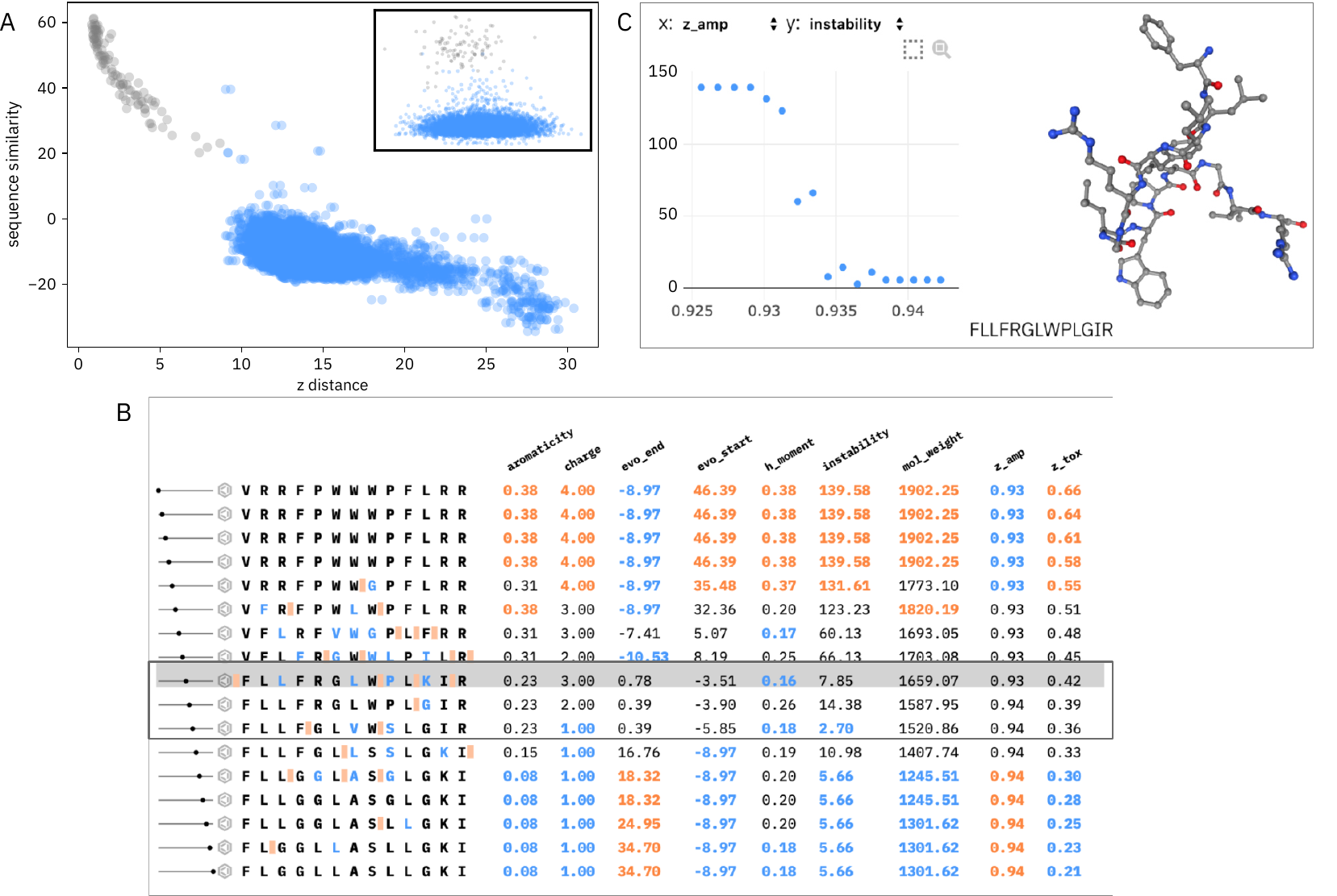}

\caption{Characteristics of the generative autoencoder latent space. (A) Relation between sequence similarity and Euclidean distance in latent $\rvz$-space, when sequences were modeled using WAE (VAE in Inset).   Darker points indicate similarity with itself (\textit{i.e.} the same exact sequence). 
Classifiers were trained either using WAE $\rvz$-space encodings ($\rvz$-) or on sequences (sequence-level). Note that the class prediction accuracy of attribute AMP using WAE z-classifiers and sequence classifiers on test data are 87.4 and 88.0, respectively. The same for toxicity attribute are 68.9 and 93.7.  
(B) Decoded sequences and their attributes during a linear interpolation between two distant sequences in the WAE latent space. 
Attributes include (1) physicochemical properties, 
  (2) sequence similarity (\texttt{evo\_start}, \texttt{evo\_end}) from  endpoint sequences,  
and (3) AMP (\texttt{z\_amp}) and Toxic (\texttt{z\_tox}) class  probabilities from $\rvz$-classifiers. 
Values in orange and blue are in the upper and lower quartile, respectively. 
Black rectangle indicates sequences with low attribute similarity to endpoint sequences.
(C) As an example of further analysis, we show the relation of AMP class probability and instability index for all candidates in the interpolation and a ball-stick rendering of a selected sequence in the path. An interactive demo of interpolations in peptide latent space is available at https://peptide-walk.mybluemix.net.
}
\label{fig:wae-latent}
\end{figure}

The advantage of training a WAE (instead of a plain VAE) on peptide sequences 
 is evident from the reported evaluation  metrics in Supplementary Information (SI) Table \ref{tab:model_results} SI. We also observed high reconstruction accuracy and diversity of generated sequences, when the WAE was trained on all peptide sequences, instead of only on AMP sequences ( Table S\ref{tab:model_results}).  Next, we analyzed the information content of the peptide WAE,  inspired by recent 
 investigations in natural language processing. By using the so-called ``probing'' methods,  it has been shown that encoded sentences can retain much linguistic information~\cite{shi2016does}.
In a similar vein, we investigated if the similarity (we use pairwise similarity which is defined by global alignment/concordance between two sequences) between sequences \cite{yu2003compositional} is captured by their encoding in the latent $\rvz$-space, as such information is known to specify the biological function and fold of peptide sequences.
  Fig. \ref{fig:wae-latent}A reveals a negative correlation (Pearson correlation coefficient = $-0.63$) between sequence similarities and  
Euclidean distances in the $\rvz$-space of the  WAE  model, suggesting that WAE intrinsically captures the sequence relationship within the peptide space. The VAE latent space fails to capture such a relation.

With the end-goal of a conditional generation of novel peptide sequences, it is crucial to ensure that the learned encoding in the $\rvz$-space retains identifiable information about functional attributes of the original sequence. 
Specifically, we investigate whether the space is \emph{linearly} separable into different attributes, such that sampling from a specific region of that space yields consistent and controlled generations. For this purpose, we trained linear classifiers for binary  (yes/no) functional attribute prediction using the $\rvz$ encodings of sequences (Fig. \ref{fig:overview-gen}B). 
Probing the $\rvz$-space modeled by the WAE  uncovers that the space is indeed linearly separable into different functional attributes, as evident from the test accuracy of binary logistic classifiers on test data: The class prediction accuracy of the attribute ``AMP" using WAE z-classifiers and
sequence-level classifiers on test data are 87.4 and 88.0, respectively 
(also see Table S\ref{tab:classifier_results}). Supplementary Table \ref{tab:AMP-compare} shows the reported accuracy of several existing AMP classification methods, as well as of our sequence-level LSTM classifier. The reported accuracy varies widely from 66.8\% for iAMP Pred \cite{meher2017predicting}, to 79\% for DBAASP-SP \cite{vishnepolsky2018predictive} that relies on local density-based sampling using physico-chemical features, to 94\%  for Witten \textit{et al.} method \cite{witten2019deep} that uses a convolutional neural net trained directly on a large  (similar to ours) corpus of peptide sequences. Our sequence-level LSTM model shows a comparable 88\% accuracy. This comparison  reveals a close performance of the $\rvz$-level classifier, when compared to classifiers reported in literature~\cite{veltri2018deep,witten2019deep,xiao2013iamp, vishnepolsky2019novo, meher2017predicting, thomas2010camp} or trained in-house that have access to the original sequences.  We emphasize that the goal of this study is not to provide a new AMP prediction method that outperforms  existing machine learning-based AMP classifiers. Rather the goal is to have a predictor trained on latent features resulting in comparable accuracy, which can be used to automatically generate new AMP candidates by conditional sampling directly from the latent space using CLaSS. It should be noted that, comparing different AMP prediction models is non-trivial, as different methods widely vary by training AMP dataset size  (\textit{e.g.} 712 for AMP Scanner v2 \cite{veltri2018deep},  3417 for iAMPpred \cite{meher2017predicting}, 2578 for CAMP \cite{thomas2010camp}, 140 for DBAASP-SP prediction \cite{vishnepolsky2019novo}, 1486 for iAMP-2L \cite{xiao2013iamp}, 4050 for \cite{witten2019deep}, and 6482 in the present study), sequence length, different definition of AMP and non-AMP, and other data curation criteria. Also, both the z-level and the sequence-level AMP classifiers used in the current study do not require any manually defined set of features, in contrast to many existing prediction tools, e.g.  \cite{vishnepolsky2019novo} and \cite{xiao2013iamp}.

On toxicity classification task, a much lower accuracy was found using models trained on latent features, when compared to similar sequence-level deep classifiers~\cite{gupta2013silico}  (also see Fig.~\ref{fig:wae-latent} and Table S\ref{tab:classifier_results}) that report accuracy as high as 90\%.  
These results imply that some attributes, such as toxicity, are more challenging to predict from the learned latent peptide representation; one possible reason can be higher class imbalance in training data (see SI). 

We also investigated the smoothness of the latent space by analyzing the sequences generated  along a linear interpolation vector in the $\rvz$-space between two distant training sequences (Fig. \ref{fig:wae-latent}B-C). Sequence similarity, functional attributes (AMP and Toxic class probabilities),  as well as several physicochemical properties including aromaticity, charge, and hydrophobic moment (indicating amphiphilicity of a helix) change smoothly during the interpolation. These results are encouraging, as the WAE latent space trained on the much larger amount of unlabeled data appears to carry significant structure in terms of functional, physicochemical, and sequence similarity. 
Figure \ref{fig:wae-latent}C also demonstrates that it is possible to identify sequence(s) during linear interpolation that is visibly different from both endpoint sequences, indicating the potential of the learned latent space for novel sequence generation.  

\subsection*{CLaSS for controlled sequence generation}
\label{sec:class}
For controlled generation, we aim to control a set of binary (yes/no) attributes of interest such as antimicrobial function and/or toxicity. We propose CLaSS - Conditional Latent (attribute) Space Sampling for this purpose. CLaSS leverages attribute classifiers directly trained on the peptide $\rvz$-space, as those can capture important attribute information (Fig. \ref{fig:wae-latent}). 
The goal  is to sample conditionally $p(\rvx | \rva_t)$ for a specified target attribute combination $\rva_t$.
This  task was approached through CLaSS (Fig. \ref{fig:overview-gen}C), which makes the assumption that attribute conditional density factors as follows:
$p(\rvx|\rva) = \int \! \mathrm{d}z \, p(\rvz|\rva) p(\rvx|\rvz)$
We sample  $p(\rvz|\rva_t)$ 
approximately using rejection sampling from models in the latent $\rvz$-space appealing to Bayes rule and $p(\rva_t|\rvz)$ modeled by the attribute classifiers (Fig. \ref{fig:overview-gen}B-C) (see SI). Since CLaSS only employs simple attribute predictor models and rejection sapling from models of $\rvz$-space, it is a simple and efficient forward-only screening method. It does not require any complex optimization over latent space, when compared to existing methods for controlled generation, \textit{e.g.} Bayesian optimization \cite{gomez2018automatic}, reinforcement learning \cite{guimaraes2017objective, popova2018deep}, or semi-supervised generative models \cite{kang2018conditional}. CLaSS is easily repurposable and embarrassingly parallelizable at the same time and does not need defining a starting point in the latent space.  

Since the Toxicity classifier trained on latent features appears weaker (Fig. \ref{fig:wae-latent}),  antimicrobial function (yes/no) was used as the sole condition for controlling the sampling from the latent peptide space. Generated antimicrobial candidates were then screened for toxicity using the sequence-level classifier during post-generation filtering. It is  noteworthy, that CLaSS does not perform a heuristic-based search (as in genetic algorithm or node-based sampling, as reported in\cite{sattarov2019novo}) on  the latent space, rather it relies on a probabilistic rejection sampling-based scheme for attribute-conditioned generation. CLaSS is also different from the local density based sampling approaches (\textit{e.g.} \cite{vishnepolsky2019novo}), as those methods rely on clustering the labeled data and then finding the cluster assignment of the test sample by using similarity search, thus is suited for a forward design task. CLaSS, in contrast, is formulated for the inverse design problem, and allows targeted generation by attribute-conditioned sampling from the latent space followed by decoding  (see Methods).


\subsection*{Features of CLaSS-generated AMPs}
\label{sec:seqanalysis} 
To check the homology of  CLaSS-generated AMP sequences with training data, we performed a BLAST sequence similarity search. We use Expect  value (E-value) for the alignment score of the actual sequences to assess sequence homology, while using other alignment metrics, such as raw alignment scores, percent identity and positive matches, gaps and coverage in alignment, to get an overall sense of sequence similarity. E-value indicates statistical (\textit{aka.} biological) significance of the match between the query and sequences from a database of a particular size. Larger E-value indicates a higher chance that the similarity between the hit and the query is merely a coincidence, \textit{i.e.} the query is not homologous or related to the hit. 
We analyzed the Expect value (E-value) for the matches with the highest alignment score.  
Typically E-values $\le$ 0.001 when querying Uniprot  $nr$ database of size $\sim$ 220 M are used to infer homology \cite{pearson2013introduction}.  Since our training database is $\sim$ 1000 times smaller than Uniprot, an E-value of $\le$ 10$^{-6}$ can be used for indicating homology. 
As shown in Table S\ref{tab:Evalue}, about 14\% of generated sequences show an E-value of $\ge$ 10, and another 36\% have an E-value $>$ 1, when considering  the match with the highest alignment score, indicating insignificant similarity between generated and training sequences. If only the alignments with score $>$ 20  are considered,  the average E-value is found to be  $>$ 2, further implying the non-homologous nature of generated sequences. Similar criteria have also been used for detecting novelty of designed short antimicrobials \cite{chen2019simulation}.  
CLaSS-generated AMPs are also more diverse, as the unique (\textit{i.e.} found only once in an ensemble of sequences) $k$-mers ($k$ = 3-6) are more abundant compared to training sequences or their fragments (Figure S\ref{tab:combined-comp}). 
These results highlight the ability of the present approach to generate short-length AMP sequences that are, on average,  novel with respect to training data, as well as diverse among themselves.  

\begin{figure}
\centering
\includegraphics[width=\textwidth]{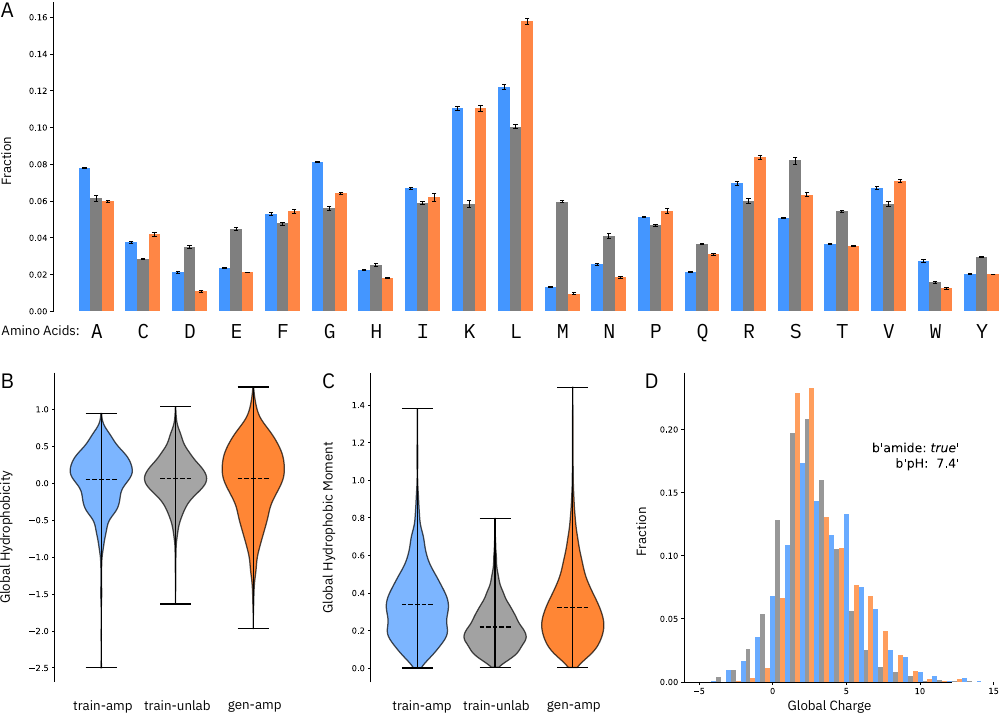}
\caption{Physico-chemical property comparison. (A) Comparison of amino acid composition, (B) global hydrophobicity, (C) hydrophobic moment (C), and (D) charge distribution of CLaSS-generated AMPs with training sequences. Mean and standard deviation  were estimated on three different sets, each consisting 3000 randomly chosen samples. Generated AMP: {\color{our_orange}orange}; training AMP: {\color{our_blue}blue}; training unlabeled: {\color{our_gray}gray}. 
}
\label{fig_biometrics}
\end{figure}

Distributions of key molecular features implicated in antimicrobial nature, such as amino acid composition, charge, hydrophobicity (H), and hydrophobic moment ($\mu$H), were compared between the training and generated AMPs, as illustrated in Fig. \ref{fig_biometrics}A-D. Additional features are reported in Figure S\ref{tab:combined-comp}. CLaSS-generated AMP sequences show distinct character: 
Specifically,  those are richer in R, L, S, Q, and C, whereas A, G, D, H,  N, and W content is reduced, in comparison to training antimicrobial sequences (Fig. \ref{fig_biometrics}A). We also present the most abundant $k$-mers ($k$=3, 4) in  Figure S\ref{tab:combined-comp}, suggesting that the most frequent 3 and 4-mers are  K and L-rich in both generated and training AMPs, the frequency being higher in generated sequences.   Generated AMPs are characterized by global net positive charge and aromaticity somewhere in between unlabeled and AMP-labeled training sequences, while the hydrophobic moment is comparable to that of known AMPs (Fig. \ref{fig_biometrics}B-D and Figure S\ref{tab:combined-comp}). These trends imply that the generated antimicrobials are still cationic and can form a putative amphiphilic $\alpha$-helix, similar to the majority of known antimicrobials.  Interestingly, they also exhibit a moderately higher hydrophobic ratio and an aliphatic index compared to training sequences (Figure S\ref{tab:combined-comp}).  These observations highlight the distinct physicochemical nature of the CLaSS-generated AMP sequences, as a result of the semi-supervised nature of our learning paradigm, that might help in their therapeutic application. For example, lower aromaticity and higher aliphatic index are known to induce better oxidation susceptibility and higher heat stability in short peptides \cite{li2016molecular}, while lower hydrophobicity is associated with reduced toxicity \cite{hawrani2008origin}. 

\subsection*{\textit{In silico} post-generation screening}
\label{sec:res-md} 

To screen the $\sim$90,000 CLaSS-generated AMP sequences, we first used an independent set of binary (yes/no) sequence-level deep neural net-based classifiers that screens for  antimicrobial function, broad-spectrum efficacy, presence of secondary structure, as well as  toxicity
 (See Figs. \ref{fig:Overview}  and  Table S\ref{tab:classifier_results}). 163 candidates passed this screening, which were then subjected to coarse-grained Molecular Dynamics (CGMD) simulations of peptide-membrane interactions. The computational efficiency of these simulations makes them an attractive choice for high-throughput and physically-inspired filtering of peptide sequences.

\begin{figure}
\centering
\includegraphics[width=1.0\textwidth]{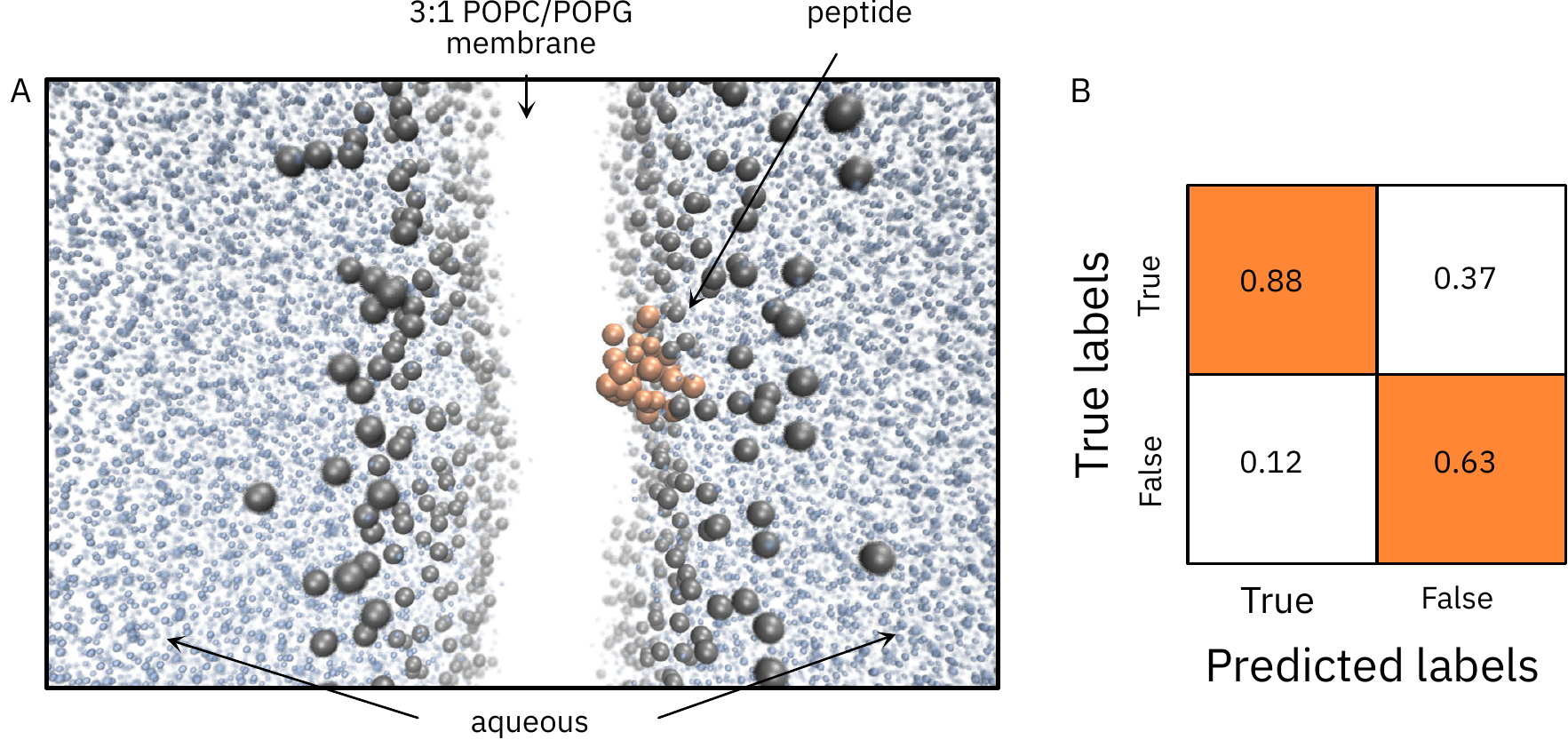}
\caption{(A) Snapshot from a coarse-grained molecular dynamics simulation of an AMP (in orange)  binding with a lipid bilayer (gray).  
(B) Confusion matrix of the simulation-based classifier that uses peptide-membrane contact variance as feature for detecting AMP sequences.
}
\label{fig:cgmd}
\end{figure}

Since there exists no standardized protocol for screening antimicrobial candidates using molecular simulations, we performed a set of control simulations of known sequences with or without antimicrobial activity. From those control runs,  we found for the first time that the variance of the number of contacts between positive residues and membrane lipids is predictive of antimicrobial activity (Supplementary Fig. \ref{fig:cgmd}): Specifically, the contact variance differentiates between high potency AMPs and non-antimicrobial sequences with a sensitivity of 88\% and specificity of 63\% (see SI). Physically, this feature can be interpreted as measuring the robust binding tendency of a peptide sequence to the model membrane. 
Therefore, we used the contact variance cutoff of $2$ for further filtering of the 163 generated AMPs that passed the classifier screening.


\subsection*{Wet lab characterization} 
\label{sec:expt}
A final set of 20 CLaSS-generated AMP sequences that  passed the contact variance-based screening mentioned above, along with their simulated and physico-chemical characteristics, are reported in  
 Tables S\ref{tab:20P-sim} and S\ref{tab:sicgmd}. Those sequences were tested in the wet lab for antimicrobial activity, as measured using minimum inhibitory concentration (MIC, lower the better) against Gram-positive S. aureus and Gram-negative E. coli (Table S\ref{tab:20P_MIC}).  
11  generated non-AMP sequences were also screened for antimicrobial activity (Table S\ref{tab:11N-MIC}). 
None of the designed non-AMP sequences showed MIC values that are low enough to be considered as antimicrobials, implying that our approach is not prone to false-negative predictions. We also speculate  that a  domain shift between the AMP test set  and generated sequences from the latent space trained on the labeled and unlabeled peptide sequences likely results in false positive prediction (18 out of 20).

\begin{figure}
\includegraphics[width=\textwidth]{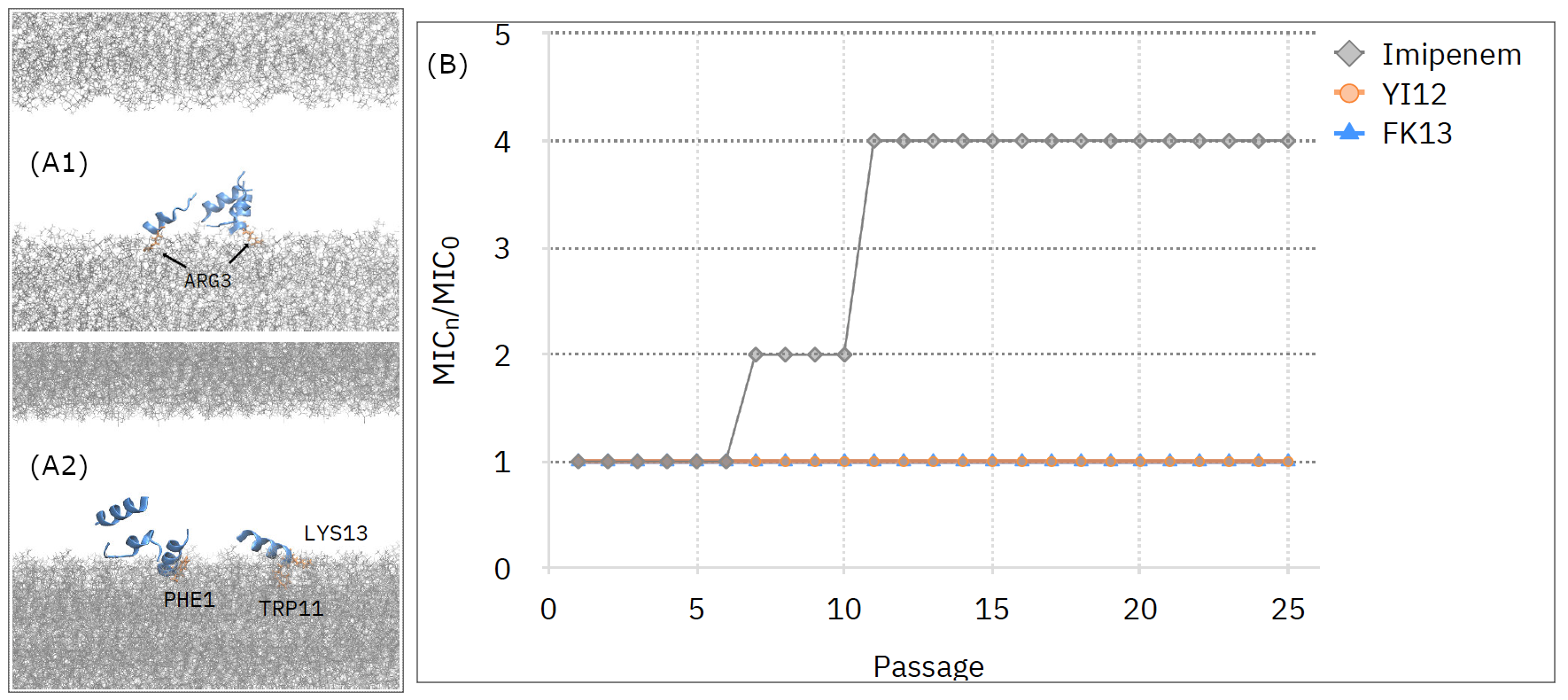}
\caption{Atomistic simulation and resistance acquisition studies. (A) Snapshot from all-atom simulation of YI12 (A1)  and FK13 (A2). Selected residues that interact with the membrane are highlighted. (B) Resistance acquisition studies of E. coli. in the presence of sub (1/2x) MIC levels of imipenem, YI12 and FK13.
}
\label{fig:6_7}
\end{figure}

Among the 20 AI-designed AMP candidates, two sequences, \textit{YLRLIRYMAKMI-CONH2} (YI12, 12 amino acids) and \textit{FPLTWLKWWKWKK-CONH2} (FK13, 13 amino acids), were identified to be the best with the lowest MIC values (Table     \ref{tab:tab-mic-compare} and Table S\ref{tab:20P_MIC}). Both peptides are positively charged and have a nonzero hydrophobic moment (Table S\ref{tab:sicgmd}), indicating their cationic and amphiphilic nature consistent with known antimicrobials. These peptides were further evaluated against the more difficult-to-treat Gram-negative \textit{P. aeruginosa},  \textit{A. baummannii}, as well as a multi-drug resistant Gram-negative \textit{K. pnuemoniae}. As listed in Fig. \ref{fig:6_7}, both YI12 and FK13 showed potent broad-spectrum antimicrobial activity with comparable MIC values. We have compared the MIC values of YI12 and FK13 with LLKKLLKKLLKK, an existing alpha helix-forming  antimicrobial peptide with excellent antimicrobial activity and selectivity reported in \cite{wiradharma2013rationally}. The MIC of LLKKLLKKLLKK against  \textit{S. aureus}, \textit{E. coli}, and \textit{P. aeruginosa} is $>$500, $>$500, and 63, respectively. MIC values of FK13 and FY12 are comparable to that of LLKKLLKKLLKK against \textit{P. aeruginosa}. However, MIC values of FK13 and YI12 are significantly lower than those of LLKKLLKKLLKK against \textit{S. aureus} and \textit{E. coli}, demonstrating greater efficacy of antimicrobial peptides discovered in this study.
We also report the results of several existing AMP prediction methods on YI12 and FK13 in Table \ref{tab:tab-mic-compare}. iAMP Pred \cite{meher2017predicting} and  CAMP-RF \cite{thomas2010camp} predict both of them as AMP, where other methods misclassify either of the two. For example, DBAASP-SP \cite{vishnepolsky2018predictive} that relies on local similarity search misclassifies FK13. Witten \textit{et al.} \cite{witten2019deep}  predicts AMP activity against \textit{S. aureus} for both sequences. The same method does not recognize FK13 as an effective antimicrobial against \textit{E. coli}.

\begin{figure}
    \captionsetup{singlelinecheck = false, justification=justified}
    \begin{subfigure}{0.33\linewidth}
        \centering
            \caption{}
            \includegraphics[scale=0.26]{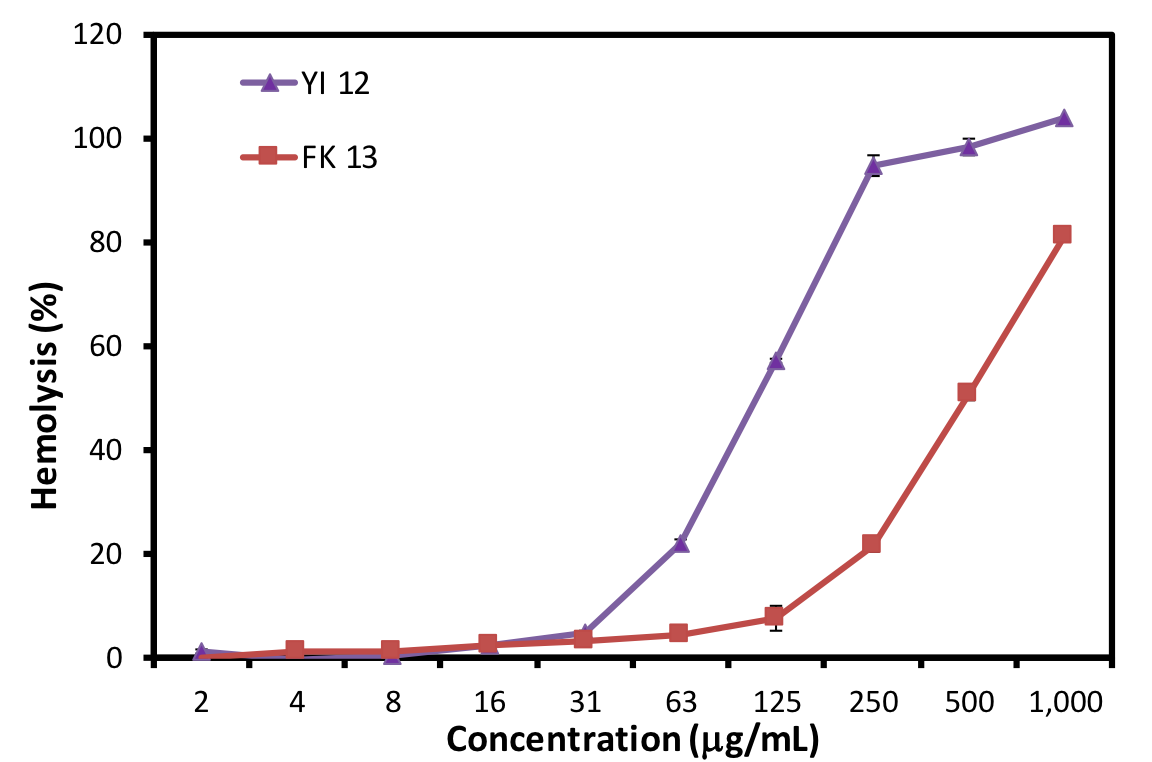}
            \label{fig:toxicity}
    \end{subfigure}
    \begin{subfigure}{0.32\linewidth}
        \centering
            \caption{}
            \includegraphics[scale=0.27]{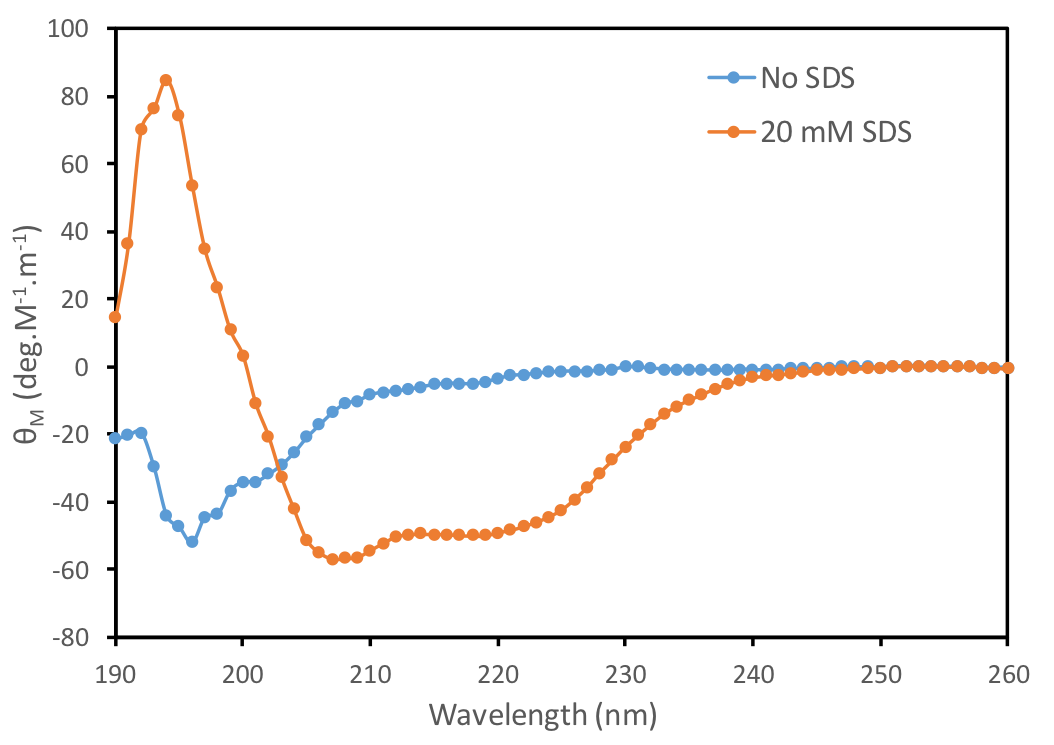}
            \label{fig:YI-CD}
    \end{subfigure}
    \begin{subfigure}{0.32\linewidth}
        \centering
            \caption{}
            \includegraphics[scale=0.27]{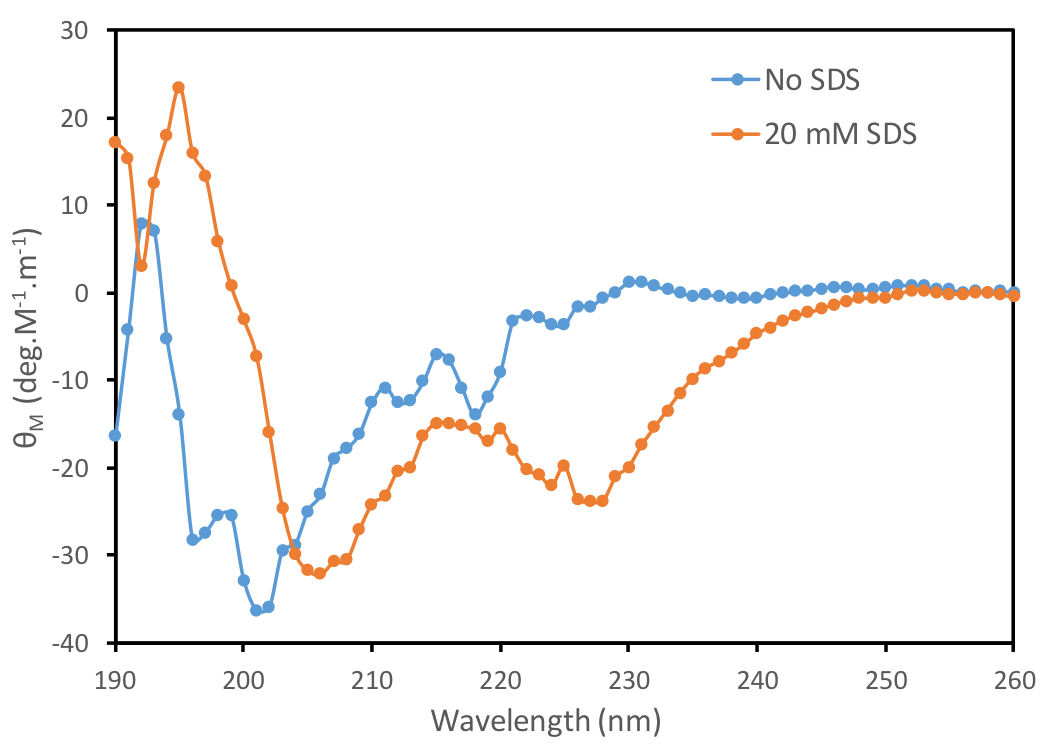}
            \label{fig:FK-CD}
        \end{subfigure}
        
        \center 
\begin{tabular}{|l|c|c|c|c|c|c|}
\toprule
\multicolumn{1}{|c|}{\textbf{Sequence}} & \textbf{Score} & \textbf{E-Value}  & \textbf{\% Coverage} & \textbf{\% Identity} & \textbf{\% Positive} & \textbf{\% Gap}\\ \hline

YLRLIRYMAKMI   & 21.88 & 1.20 & 66.67 & 75.00 & 75.00 & 75.00 \\ \hline
FPLTWLKWWKWKK  & 33.30 & 4e-05 & 84.61 & 73.00 & 73.00 & 9.00\\ 
\bottomrule
\end{tabular}

\caption{Percentage of hemolysis of rat red blood cells as a function of peptide concentration. (B) and (C) show CD Spectra of YI12 and FK13 peptide, respectively, at 0.5 mg/ml concentration in DI water and presence of 20 mM SDS buffer. Both YI12 and FK13 showed a random coil-like structure in the absence of SDS. When SDS was present, both sequences form $\alpha$-helical structure (evident from the 208 nm and 222 nm peaks). (D) BLAST search results (alignment score, E-value, percentage of alignment coverage, percentage of identity, percentage of positive matches or similarity,  percentage of alignment gap) against full training data for YI12 and FK13.}
\label{fig:cd_n_tox}
\end{figure}

We further performed \textit{in vitro} and \textit{in vivo} testing for toxicity. 
 Based on  activity measure at 50\% hemolysis ($HC_{50}$) and lethal dose ($LD_{50}$) toxicity values (Table \ref{tab:tab-mic-compare} and Fig. S\ref{fig:cd_n_tox}), both peptides appear biocompatible (as the $HC_{50}$ and $LD_{50}$ values are much higher than MIC values), FK13 being more biocompatible than YI12. More importantly, the $LD_{50}$ values of both peptides compare favorably with that of polymyxin B (20.5 mg/kg) \cite{rifkind1967prevention}, which is a clinically used antimicrobial drug for treatment of antibiotic-resistant Gram-negative bacterial infection.

\subsection*{Sequence similarity analyses}
\label{sec:novelty}
\begin{figure}
\centering
\includegraphics[width=\textwidth]{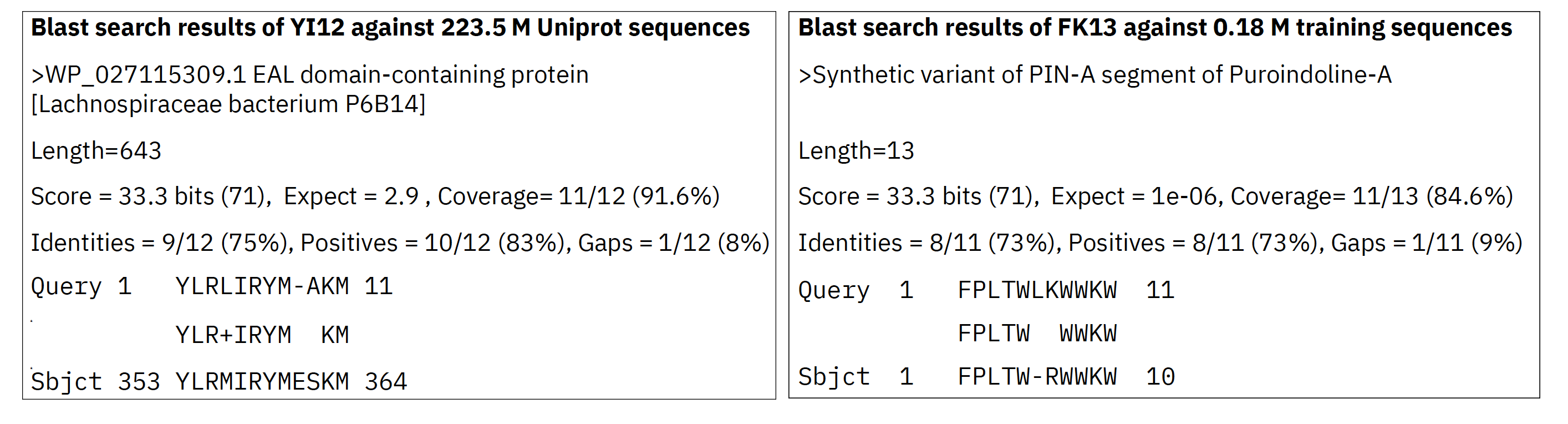}
\caption{BLAST search results of YI12 and FK13.}. 
\label{fig:blast_result}
\end{figure}

To investigate the similarity of YI12 and FK13 with respect to training sequences, we analyzed the alignment scores  returned by the BLAST homology search in detail (Fig. S\ref{fig:blast_result} and  Fig. S\ref{fig:cd_n_tox}), in line with earlier works \cite{ronvcevic2018parallel, chen2019simulation}. Scoring metrics include raw alignment score, E-value, percentage of alignment coverage, percentage of identity, percentage of positive matches or similarity,  and percentage of alignment gap (indicating the presence of additional amino acids). BLAST searching with an E-value threshold of 10 against the training database did not reveal any match for YI12, suggesting that there exists no statistically significant match of YI12. Therefore, we further searched for related sequences of  YI12  in the much larger Uniprot database consisting of $\sim$ 223.5 M non-redundant sequences, only a fraction of which was included in our model training. YI12 shows an E-value of 2.9 to its closest match, 

which is an 11 residue segment from the bacterial EAL domain-containing protein (Fig. S\ref{fig:blast_result}). This result suggests that  YI12  shares low similarity, even when all protein sequences in Uniprot are considered. We also performed a BLAST search of YI12 against the PATSEQ database that contains $\sim$ 65.5 M patented peptides and still received a minimum E-value of 1.66. The sequence nearest to YI12 from PATSEQ is an eight amino acid long segment from a 79 amino acid long human protein, which has with 87.5\% similarity and only 66.7\% coverage, further confirming YI12's low similarity to known sequences.

FK13 shows less than 75\% identity, a gap in the alignment, and  85\% query coverage to its closest match, in the training database, implying  FK13 also shares low similarity  to training sequences (Fig. S\ref{fig:blast_result}).  YI12 is more novel than FK13, though. The closest match of FK13 in the training database is a synthetic variant of a 13 amino acid long bactericidal domain (PuroA: FPVTWRWWKWWKG) of  Puroindoline-A protein from wheat endosperm.  The antimicrobial and hemolysis activities of FK13 are close to those reported for PuroA \cite{jing2003conformation, haney2013mechanism}. Nevertheless, FK13 is significantly different from PuroA; FK13  is K-rich and low in W-content, resulting in lower Grand Average of Hydropathy (GRAVY) score ($-0.854$ \textit{vs.} $-0.962$), higher aliphatic index ($60.0$ \textit{vs.} $22.3$), and  lower instability index ($15.45$ \textit{vs.} $58.30$), all together indicative of higher peptide stability. In fact, lower W-content was found beneficial for stabilizing of FK13 during wet-lab experiments, since Tryptophan (W) is susceptible to oxidation in air. Lower W-content has also been implicated in improving  \textit{in vivo} peptide stability \cite{mathur2018silico}. 
Taken together, these results illustrate that CLaSS on latent peptide space modeled by the WAE is able to generate novel and optimal antimicrobial sequences by efficiently learning the complicated sequence-function relationship in peptides and exploiting that knowledge for controlled exploration. When combined with subsequent \textit{in silico} screening,  novel and optimal lead candidates with experimentally confirmed high broad-spectrum efficacy and selectivity are identified at a success rate of 10\%. The whole cycle (from database curation to wet lab confirmation) took 48 days in total and a single iteration (Fig. \ref{fig:Overview}).  

 \subsection*{Structural and mechanistic analyses}
 \label{sec:str}
 
 We performed all-atom explicit water simulations (see SI) of these two sequences in the presence of a lipid membrane starting from an $\alpha$-helical structure. Different membrane binding mechanisms were observed for the two sequences, as illustrated in Fig.~\ref{fig:6_7}A. YI12 embeds into the membrane by using positively charged N-terminal Arginine (R) residues. While FK13 embeds either with N-terminal Phenylalanine (F) or with C-terminal Tryptophan (W) and  Lysine (K). These results provide mechanistic insights into different modes of action adopted by YI12 and FK13 during the early stages of membrane interaction.
 
 Peptides were further experimentally characterized using CD spectroscopy (see SI). 
 Both YI12 and FK13 showed random coil-like structure in water, but formed an $\alpha$-helix in 20\% SDS buffer (Fig. S\ref{fig:cd_n_tox}). Structure classifier predictions (see SI) and all-atom simulations   are consistent with the CD resylts. From  CD spectra, $\alpha$-helicity of YI12 appears stronger than that of FK13, in line with its stronger hydrophobic moment (Table S\ref{tab:sicgmd}). In summary,  physicochemical analyses and CD spectroscopy together suggest that cationic nature and amphiphilic helical topology are the underlying factors inducing antimicrobial nature in YI12 and FK13.
 

 To provide insight onto the mechanism of action underlying the antimicrobial activity of YI12 and FK13, we conducted agar plate assay and found that both peptides YI12 and FK13 are bactericidal. There was a 99.9\% reduction of colonies at 2 x MIC.
 
\begin{figure}
\centering
\includegraphics[width=\textwidth]{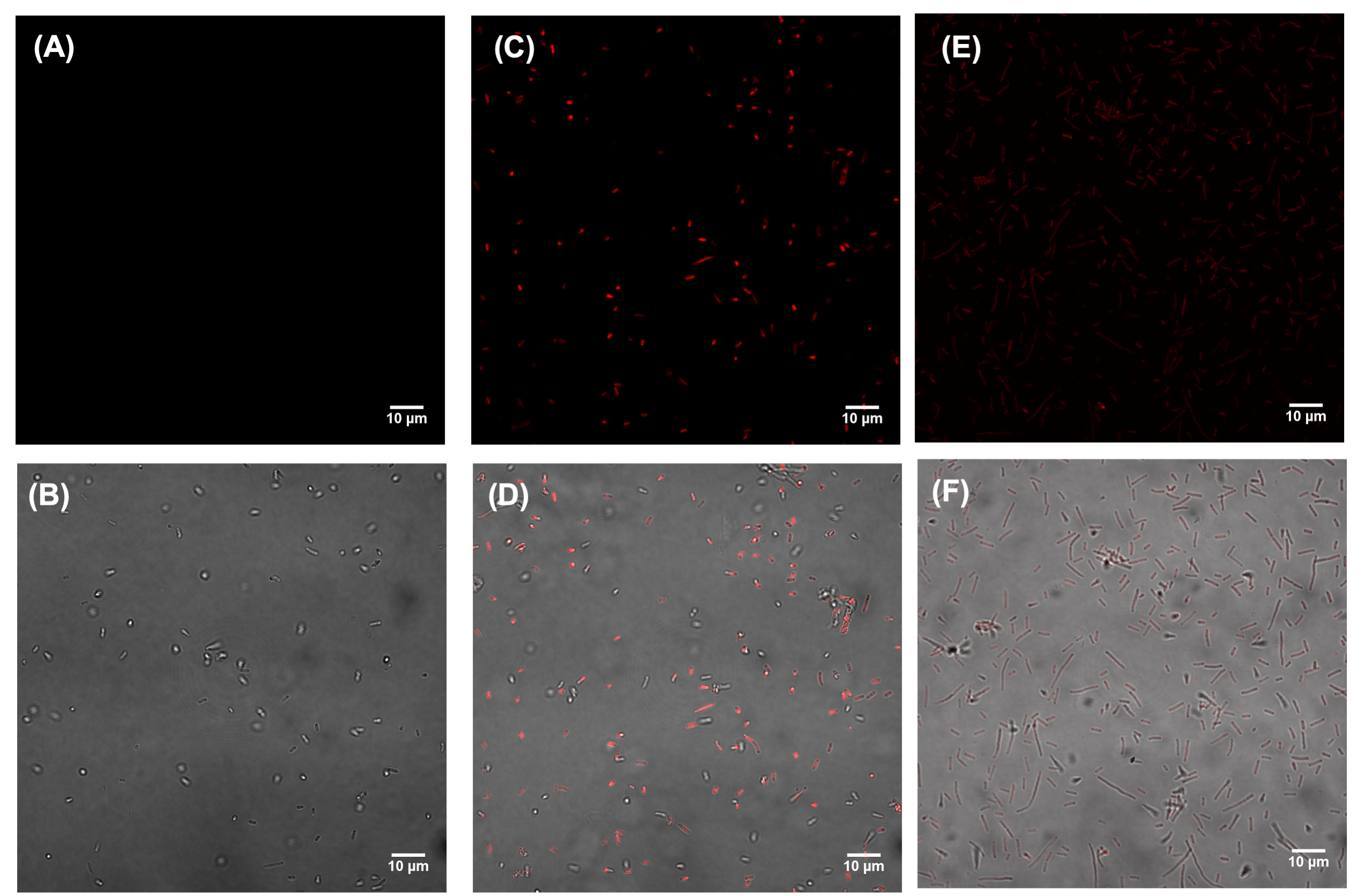}
\caption{Confocal images of \textit{E. coli} (A-B) without any treatment, (C-D) treated with 2xMIC FK13 and (E-F) treated with 2xMIC polymyxin B for 2 hours. Red colour denotes the propidium iodide. Images A, C and E are the dark field image and B, D and F are the bright field image. All scale bars are 10 $\mu$m.}. 
\label{fig:confocal}
\end{figure}

 Since $\alpha$-helical peptides like  YI12 and FK13 are known to disrupt membranes by leaky pore formation ~\cite{kumar2018antimicrobial,guha2019mechanistic}, we performed  live-imaging with confocal fluorescence microscopy for FK13 against \textit{E. coli} (Figure \ref{fig:confocal}). The results were compared with that of polymyxin B, that is one of a group of basic polypeptide antibiotics derived from \textit{B polymyxa}.   After 2 hours of  incubating  E. coli with polymyxin B or with FK13 at 2 x MIC,   confocal imaging was performed.  Emergence of fluorescence, as shown in Figure \ref{fig:confocal}, confirmed that red propidium iodide (PI) has entered the bacterial cell and interacted with bacterial DNA in the presence of either FK13 or polymyxin B.   This finding implies that both polymyxin and FK13 induce pore formation on the bacteria membrane and allow the PI dye to enter the bacteria. Without the pore formation, the PI dye will not be able to enter the bacteria.
 
 \subsection*{Resistance analyses}
 Finally, we  performed resistance acquisition studies of E. coli. in the presence of  imipenem, an intravenous $\beta$-lactam antibiotic,  YI12 or FK13 at concentrations of sub MIC levels. Results shown in Figure \ref{fig:6_7}B confirm that both YI12 and FK13 do not induce resistance after 25 passages, while E. coli has begun to develop resistance to the antibiotic Imipenem after just 6 passages. We have also investigated the efficacy of these peptides against polymyxin B resistant \textit{K. pneumoniae}, a strain that is resistant to polymyxin B, a beta-lactum antibiotic, an antibiotic of last resort. Table \ref{tab:tab-mic-compare} shows the MIC values of   YI12, FK13, and polymyxin B, revealing no  MIC increase for either of the two discovered peptides, when compared to the MIC against the MDR K. pneumoniae stain (from ATCC). In contrast, the MIC of polymyxin B is 2 $\mu$g/ml against the same MDR K. pneumoniae stain (from ATCC), but increases to  $>$125 $\mu$g/ml once the \textit{K. pneumoniae} became resistant to polymyxin B. The MIC value of YI12 and FK13 is still lower than that of polymyxin B in the polymyxin B resistant strain, indicating that the resistance to polymyxin B  is not seen towards YI12 and FK13.  Taken together, these results indicate that YI12 and FK13 hold therapeutic potential for treating resistant strains and therefore demand further investigation.


\section*{Discussion and Conclusions}
\label{sec:conclusion}
Learning implicit interaction rule(s) of complex molecular systems is a  major goal of artificial intelligence (AI) research. 
This direction is critical for designing new molecules/materials with specific structural and/or functional requirements, one of the most anticipated and acutely needed applications. 
Antimicrobial peptides considered here represent an archetypal system for molecular discovery problems. They exhibit a near-infinite and mostly unexplored chemical repertoire,
a well-defined chemical palette (natural amino acids), as well as potentially conflicting or opposing design objectives, and is of high importance due to the global increase in antibiotic resistance and a depleted antibiotic discovery pipeline. Recent work has shown that deep learning can be used to help screen libraries of existing chemicals for antibiotic properties \cite{stokes2020deep}. A number of recent studies have also used AI methods for  design of antimicrobial peptides and provided experimental validation  \cite{loose2006linguistic,  nagarajan2018computational, maccari2013antimicrobial, porto2018silico, fjell2011optimization, vishnepolsky2018predictive, nagarajan2019omega76}. However, to our knowledge, the present work provides for the first time a fully automated computational framework that combines  controllable generative modeling, deep learning, and physics-driven learning for \textit{de novo} design of broad-spectrum potent and selective AMP sequences and experimentally validates them for broad-spectrum efficacy and toxicity. Further, the discovered peptides show high efficacy against a strain that is resistant to an antibiotic of last resort, as well as mitigate drug resistance onset.
Wet lab results confirmed the efficiency of the proposed approach for designing novel and optimized sequences with a very modest number of candidate compounds synthesized and tested. The present design approach in this proof-of-concept study yielded a 10\% success rate and a rapid turnaround of
48 days, highlighting the importance of combining AI-drive computational strategies with experiments to achieve more effective drug candidates.
The generative modeling approach presented here can be tuned for not only generating novel candidates, but also for designing novel combination therapies and antibiotic adjuvants, to further advance antibiotic treatments.

Since CLaSS is a generic approach, it is suitable for a variety of controlled generation tasks and can handle multiple controls simultaneously.  The method is simple to implement, fast, efficient, and scalable, as it does not require any optimization over the latent space. CLaSS has additional advantages regarding repurposability, as adding a new constraint requires a simple predictor training.  
Therefore,  future directions of this work will explore the effect of 
additional relevant constraints, such as the induced resistance, efficacy in animal models of infection, and fine-grained strain-specificity, on the  designed AMPs using the approach presented here. Extending CLaSS application to other controlled molecule design tasks, such as target-specific and selective drug-like small molecule generation is also underway \cite{chenthamarakshan2020target}.
Finally,  the AI models will be further optimized in an iterative manner 
by using the feedback from simulations and/or experiments in an active learning framework.  

\clearpage

\subsection*{Online Methods}
\subsubsection*{Generative Autoencoders}
To learn meaningful continuous latent representations from sequences without supervision, the Variational Autoencoder (VAE) \cite{kingma2013auto} family has emerged as a principled and successful method. 
The data distribution $p(\rvx)$ over samples $\rvx$ is represented as the marginal of a joint distribution $p(\rvx, \rvz)$ that factors out as $p(\rvz) p_\theta(\rvx|\rvz)$.
The prior $p(\rvz)$ is a simple smooth distribution, while $p_\theta(\rvx | \rvz)$ is the decoder that maps a point in latent $\rvz$-space to a distribution in $\rvx$ data space.
The exact inference of the hidden variable $\rvz$ for a given input $\rvx$ would require integration over the full latent space: $p(\rvz | \rvx) = \frac{p(\rvz) p_\theta(\rvx|\rvz)}{\int d\rvz p(\rvz) p_\theta(\rvx|\rvz)}$.
To avoid this computational burden, the inference is approximated through an inference neural network or encoder $q_\phi(\rvz|\rvx)$.
Our implementation follows \cite{bowman2015large}, where both encoder and decoder are single-layer LSTM recurrent neural networks \cite{hochreiter1997long},
and the encoder specifies a diagonal Gaussian distribution, i.e. $q_\phi(z|x) = N(z; \mu(x), \Sigma(x))$ (Fig. \ref{fig:overview-gen}).

The basis for auto-encoder training is optimization of an objective consisting of the sum of a reconstruction loss and a regularization constraint loss term: $\gL(\theta, \phi) = \gL_{rec}(\theta, \phi) + \gL_c(\phi)$.
In the standard VAE objective \cite{kingma2013auto}, reconstruction loss $\gL_{rec}(\theta, \phi)$ is based on the negative log likelihood of the training sample, 
and the constraint $\gL_c(\phi)$ uses $D_{KL}$, the Kullback-Leibler divergence:
\[
     \mathcal{L}_{\text{VAE}}(\theta, \phi) = 
    \E_{q_\phi(z|x)}[\log p_\theta(x|z)]
    - D_{KL}(q_\phi(z|x) || p(z))
\]
for a single sample.
This exact objective is derived from a lower bound on the data likelihood; hence this objective is called the ELBO (Evidence Lower Bound). With the standard VAE, we observed the same posterior collapse as detailed for natural language in the literature \cite{bowman2015generating},
meaning $q(z|x) \approx p(z)$ such that no meaningful information is encoded in $z$ space.
Further extensions include $\beta$-VAE that adds a multiplier ``weight'' hyperparameter $\beta$ on the regularization term, and $\delta$-VAE that encourages the $D_{KL}$ term to be close to a nonzero $\delta$, \textit{etc.}, to tackle the issue of posterior collapse. However, finding the right setting that serves as a workaround for the posterior collapse is tricky within these VAE variants. 

Therefore, many variations within the VAE family have been recently proposed, such as Wasserstein Autoencoder (WAE) \cite{tolstikhin2017wasserstein, bahuleyan2018probabilistic} and Adversarial Autoencoder (AAE) \cite{makhzani2015adversarial}.

WAE factors an optimal transport plan through the encoder-decoder pair,
on the constraint that marginal posterior $q_\phi(z) = \E_{x \sim p(x)} q_\phi(z|x)$
equals a prior distribution, i.e. $q_\phi(z)=p(z)$.
This is relaxed to an objective similar to $\mathcal{L}_{\text{VAE}}$ above. However, 
in the WAE objective \cite{tolstikhin2017wasserstein}, instead of each individual $q_\phi(\rvz | \rvx)$, the marginal posterior $q_\phi(\rvz) = \E_\rvx [q_\phi(\rvz|\rvx)]$ is constrained to be close to the prior $p(\rvz)$.
We enforce the constraint by penalizing maximum mean discrepancy \cite{gretton2007kernel} with random features approximation of the radial basis function \cite{rahimi2007random}:
$\gL_c(\phi) = \text{MMD}(q_\phi(\rvz), p(\rvz))$.
The total objective for WAE is $\gL = \gL_{rec} + \gL_c$ where we use the reconstruction loss $\gL_{rec} = -\E_{q_\phi(\rvz|\rvx)}[\log p_\theta(\rvx|\rvz)]$. In WAE training with maximum mean discrepancy (MMD) or with a discriminator, we found a benefit of regularizing the encoder variance as in the literature~\cite{rubenstein2018latent,bahuleyan2018probabilistic}. For MMD, we used a random features approximation of the Gaussian kernel \cite{rahimi2007random}.

Details of autoencoder architecture and training, as well as an experimental comparison between different auto-encoder variations tested in this study, can be found in Supplementary Material sections \ref{si:models}, \ref{method:training} and \ref{si:vaewaeaae}. Python codes for training peptide autoencoders are available via github at https://github.com/IBM/controlled-peptide-generation.

\subsubsection*{CLaSS - Conditional Latent (Attribute) Space Sampling}
\label{si:lcs}
\begin{figure}
\centering
\includegraphics[width=\textwidth]{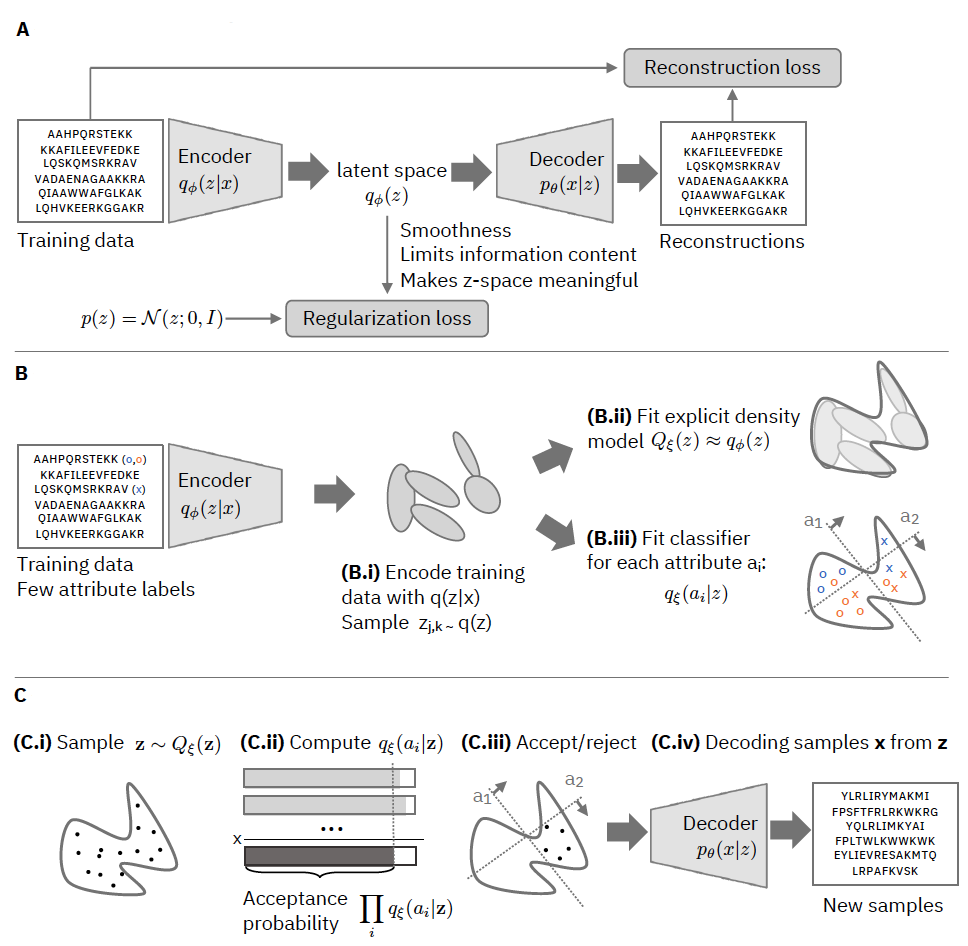}
\caption{Phases of attribute-controlled peptide sequence generation. (A) Training a generative Autoencoder (AE) model on peptide sequences (AE training in Fig. 1), (B) Mapping sparse peptide attributes to the model's latent $\rvz$-space and constructing the density model of the $\rvz$-space (Autoencoder Evaluation in Fig. 1), and (C) Sampling from the $\rvz$-space using our CLaSS method (Controlled Generation in Fig. 1).  }
\label{fig:overview-gen}
\end{figure}

We propose Conditional Latent (attribute) Space Sampling, CLaSS, a simple but efficient method to sample from the targeted region of the latent space from an auto-encoder, which was trained in an unsupervised manner (Fig. \ref{fig:overview-gen}).

\textbf{Density Modeling in Latent Space}
We assume a latent variable model (e.g., Autoencoder) that has been trained in an unsupervised manner to meet the evaluation criteria outlined in (see SI). 
 All training data $\rvx_j$ are then encoded in latent space: $\rvz_{j,k} \sim q_\phi(\rvz | \rvx_j)$.
These $\rvz_{j,k}$ are used to fit an explicit density model $Q_\xi(\rvz)$ to approximate marginal posterior $q_\phi(\rvz)$, and a classifier model $q_\xi(a_i|\rvz)$ for attribute $a_i$ to approximate the probability $p(a_i|\rvx)$.
The motivation for fitting a $Q_\xi(\rvz)$ is in order to sample from $Q_\xi$ rather than $p(\rvz)$, since at the end of training the discrepancy between $q_\phi(\rvz)$ and $p(\rvz)$ can be significant.

Although any explicit density estimator could be used for $Q_\xi(\rvz)$,
here we consider Gaussian mixture density models and evaluate negative log-likelihood on a held-out set to determine the optimal complexity.
We find 100 components and untied diagonal covariance matrices to be optimal, giving a held-out log likelihood of $105.1$.
To fit $Q_\xi$, we use K=10 random samples from the encoding distribution of the training data,  $\rvz_{j,k} \sim q_\phi(\rvz | \rvx_j) = \mathcal{N}(\mu(\rvx_j), \sigma(\rvx_j))$, with $k=1 \dots K$.

 Independent simple linear attribute classifiers $q_\xi(a_i | \rvz)$ are then fitted per attribute.
For each attribute $a_i$, the procedure consists of: 
(1) collecting dataset with all labeled samples for this attribute $(\rvx_j, a_i)$, 
(2) encoding the labeled data as before, $\rvz_{j,k} \sim q_\phi(\rvz | \rvx_j)$, 
(3) fitting $\xi$, the parameters of logistic regression classifier $q_\xi(a_i | \rvz)$ with inverse regularization strength $C=1.0$ and 300 lbfgs iterations.

\textbf{Rejection Sampling for Attribute-Conditioned Generation}
Let us formalize that there are $n$ different (and possibly independent) binary attributes of interest $\rva \in \{0,1\}^n = [a_1, a_2, \dots, a_n]$, each attribute is only available (labeled) for a small and possibly disjoint subset of the dataset. Since functional annotation of peptide sequences is expensive, current databases typically represent a small  ($\approx 100-10000$)  subset of the unlabeled corpus.
We posit that all plausible datapoints have those attributes, albeit mostly without label annotation. Therefore, the data distribution implicitly is generated as $p(\rvx) = \E_{\rva \sim p(\rva)} [ p(\rvx | \rva)]$, where the distribution over the (potentially huge) discrete set of attribute combinations $p(\rva)$ is integrated out, and for each attribute combination the set of possible sequences is specified as $p(\rvx | \rva)$.
As our aim is to sample novel sequences $\rvx \sim p(\rvx | \rva)$, for a desired attribute combination $\rva = [a_1, \dots, a_n]$, 
we are now able to approach this task through conditional sampling in latent space:
\begin{align}
  p(\rvx|\rva) & = \int \! \mathrm{d}z \, p(\rvz|\rva) p(\rvx|\rvz) \\
  & \approx \int \! \mathrm{d}z \, \hat{p}_\xi(\rvz|\rva) p_\theta(\rvx|\rvz)
\end{align}
Where $\hat{p}_\xi(\rvz|\rva)$ will not be approximated explicitly, rather we will use rejection sampling using the models $Q_\xi(\rvz)$ and $q_\xi(a_i | \rvz)$ to approximate samples from $p(\rvz | \rva)$.

To approach this, we first use Bayes' rule and the conditional independence of the attributes $a_i$  conditioned on $\rvz$, since we assume the latent variable captures all information to model the attributes: $a_i \perp a_j | \rvz$ (\textit{i.e.} two attributes $a_i$ and $a_j$ are independent when conditioned on $\rvz$)
\begin{align}
p(\rvz|\rva) 
 &= \frac{p(\rva | \rvz) q_\phi(\rvz)} {p(\rva)} \\
 & = \frac{ q_\phi(\rvz) \prod_i p(a_i | \rvz)} {p(\rva)}
\end{align}

This approximation is introduced to $\hat{p}_\xi(\rvz|\rva)$,
using the models $Q_\xi$ and $q_\xi$ above:
\begin{align}
\hat{p}_\xi(\rvz|\rva) &= \frac{Q_\xi(\rvz) \prod_i q_\xi (a_i | \rvz)}{q_\xi(\rva)}
\label{eq:rejsamplingbase}
\end{align}

The denominator $q_\xi(\rva)$ in Eq. (\ref{eq:rejsamplingbase}) could be estimated by approximating the expectation $q_\xi(\rva) = \E_{Q_\xi(\rvz)} q_\xi(\rva | \rvz) \approx \frac{1}{N} \sum_{\rvz_j \sim Q_\xi(\rvz)}^N q_\xi(\rva | \rvz)$.
However, the denominator is not needed \textit{a priori} in our rejection sampling scheme,
in contrast, $q_\xi(\rva)$ will naturally appear as the rejection rate of samples from the proposal distribution (see below).

For rejection sampling distribution with pdf $f(\rvz)$, we need a proposal distribution $g(\rvz)$ and a constant $M$, such that $f(\rvz) \leq M g(\rvz)$ for all $\rvz$, i.e. $M g(\rvz)$ envelopes $f(\rvz)$.
We draw samples from $g(\rvz)$ and accept the sample with probability $\frac{f(\rvz)}{M g(\rvz)}  \leq  1$.

In the above, to sample from Eq. (\ref{eq:rejsamplingbase}), we consider $\rva$ to be constant.
We  perform rejection sampling through the proposal distribution: $g(\rvz) = Q_\xi(\rvz)$ that can be directly sampled.
Now set $M=1/q_\xi(\rva)$ so $M g(\rvz) = Q_\xi(z) / q_\xi (\rva)$,
while our pdf to sample from is $f(\rvz) = Q_\xi(\rvz) \prod_i q_\xi (a_i | \rvz) / q_\xi(\rva)$.
Therefore, we accept the sample from $Q_\xi(\rvz)$ with probability
\[
\frac{f(\rvz)}{M g(\rvz)} = \prod_i q_\xi (a_i | \rvz)  \leq 1
\]
The inequality trivially follows from the product of normalized probabilities.
The acceptance rate is $1/M = q_\xi(\rva)$.
Intuitively, the acceptance probability is equal to the product of the classifier's scores, while sampling from explicit density $Q_\xi(z)$. 
In order to accept any samples, we need a region in $\rvz$ space to exist where $Q_\xi(\rvz) > 0$ and the classifiers assign a nonzero probability to all desired attributes, \textit{i.e.} the combination of attributes has to be realizable in $\rvz$-space.

\clearpage

\newcommand{\specialcell}[2][c]{%
\begin{tabular}[#1]{@{}c@{}}#2\end{tabular}}
  
\begin{table}
    \centering
    \begin{tabular}{c|c|c|c|c|c|c|c|c}
        \toprule
         Sequence & \specialcell{SA} & \specialcell{EC} & \specialcell{PA } & \specialcell{AB} & \specialcell{MDR-KP
} & \specialcell{polyR-KP} &  HC$_{50}$ & LD$_{50}$ \\ 
        \hline
        YI12 & 7.80 & 31.25 & 125.00 & 15.60 & 31.25  & 31 & 125 & 182 \\
        FK13 & 15.60 & 31.25 & 62.50 & 31.25 & 15.60  & 16 & 500 & 158 \\
        
         \bottomrule
    \end{tabular}
    \caption{Antimicrobial activity and toxicity values of YI12 and FK13, two best CLaSS-designed AMPs.
    MIC values against diverse strains, including SA (S. aureus), EC (E. coli), PA (P. aeruginosa), AB (A. baumannii), one multidrug-resistant (MDR) \textit{K. pneumonia} (MDR-KP), and one polymyxin B resistant \textit{K. pneumonia} (polyR-KP). Hemolytic activity measured at 50\% hemolysis (HC$_{50}$) using rat red blood cells,  and lethal dose toxicity (LD$_{50}$) values are for Balb/C mice. All reported MIC and HC$_{50}$ values are in $\mu$g/mL, while unit for LD$_{50}$ is mg/kg. As a baseline, MIC value of polymyxin B against polyR-KP is $>$125 $\mu$g/mL.}
    \label{tab:tab-mic-compare}
\end{table}

\clearpage

\section*{Acknowledgments}
We acknowledge Youssef Mroueh and Kush Varshney for insightful discussions. We also thank Oscar Chang, Elham Khabiri, and Matt Riemer for help with the initial phase of the work. We would like to acknowledge David Cox, Yuhai Tu, Pablo Meyer Rojas, and Mattia Rigotti for providing valuable feedback on the manuscript. F.C. thanks Patrick Simcock for sharing knowledge. We thank anonymous reviewers for constructive feedback that significantly improved quality of this manuscript.

\section*{Author Information}
Correspondence and requests for materials should be addressed to 
Payel Das~(email: daspa@us.ibm.com).

\textbf{Author Contributions:} P.D., J.C., and A.M. conceived the project; P.D. designed and managed the project; P.D and T.S. designed and implemented the sequence generation and screening framework and algorithm with help from K.W., I.P., and S.G.; Autoencoder experiments were run and analyzed by P.D., T.S., K.W., I.P., P.C., V.C., and C.D.S.; Generated sequences were analyzed in silico by P.D., T.S., H.S., and K.W.; F.C. with help from P.D. and J.C. designed, performed and analyzed the molecular dynamics simulations; Y.Y.Y., J.T., J.H.  performed and analyzed the wet lab experiments; H.S. created final figures; All authors wrote the paper. 

\section{Ethics Statement}
The authors declare that they have no competing financial interests. 

\section{Data and materials availability:} A full description of the model and method
is available in the supplementary materials. Data and code will be available upon request and is also accessible via github at https://github.com/IBM/controlled-peptide-generation/tree/master/models.  

\section{Supplementary Information}
\textbf{Supplementary Information}
Supplementary tables, figures, descriptions of the dataset and details of models and methods.

\textbf{Reporting Summary}
 Further information on research design is available in the Nature Research Reporting Summary linked to this article.

\clearpage

\bibliographystyle{Science}
\bibliography{scibib}

\begin{thebibliography}{100}

\bibitem{dimasi2016innovation}
J.~A. DiMasi, H.~G. Grabowski, R.~W. Hansen, {\it Journal of health
  economics\/} {\bf 47}, 20 (2016).

\bibitem{desselle2017institutional}
M.~R. Desselle, {\it et~al.\/}, {\it Future Science OA\/} {\bf 3}, FSO171
  (2017).

\bibitem{UN_2019}
U.~Nations, No time to wait: {Securing} the future from drug-resistant
  infections, {\it Tech. rep.\/}, United Nations (2019).

\bibitem{oneill_2016}
J.~O'Neill, Tackling {Drug}-{Resistant} {Infections} {Globally}: final report
  and recommendations, {\it Tech. rep.\/}, The Review on Antimicrobial
  Resistance (2016).

\bibitem{WHO_2019}
W.~H. Organization, {\it et~al.\/}, 2019 {Antibacterial} agents in clinical
  development, {\it Tech. rep.\/}, WHO (2019).

\bibitem{Powers2003}
J.-P.~S. Powers, R.~E. Hancock, {\it Peptides\/} {\bf 24}, 1681 (2003).

\bibitem{Mahlapuu2016}
M.~Mahlapuu, J.~H{\aa}kansson, L.~Ringstad, C.~Bj\"{o}rn, {\it Frontiers in
  Cellular and Infection Microbiology\/} {\bf 6}, 194 (2016).

\bibitem{chen2019simulation}
C.~H. Chen, {\it et~al.\/}, {\it Journal of the American Chemical Society\/}
  {\bf 141}, 4839 (2019).

\bibitem{torres2018structure}
M.~D. Torres, {\it et~al.\/}, {\it Communications biology\/} {\bf 1}, 1 (2018).

\bibitem{tucker2018discovery}
A.~T. Tucker, {\it et~al.\/}, {\it Cell\/} {\bf 172}, 618 (2018).

\bibitem{field2013saturation}
D.~Field, {\it et~al.\/}, {\it Microbial biotechnology\/} {\bf 6}, 564 (2013).

\bibitem{fjell2012designing}
C.~D. Fjell, J.~A. Hiss, R.~E. Hancock, G.~Schneider, {\it Nature reviews Drug
  discovery\/} {\bf 11}, 37 (2012).

\bibitem{li2017membrane}
J.~Li, {\it et~al.\/}, {\it Frontiers in neuroscience\/} {\bf 11}, 73 (2017).

\bibitem{cardoso2020computer}
M.~H. Cardoso, {\it et~al.\/}, {\it Frontiers in Microbiology\/} {\bf 10}, 3097
  (2020).

\bibitem{jenssen2008qsar}
H.~Jenssen, C.~D. Fjell, A.~Cherkasov, R.~E. Hancock, {\it J. Pept. Sci\/} {\bf
  14}, 110 (2008).

\bibitem{vishnepolsky2019novo}
B.~Vishnepolsky, {\it et~al.\/}, {\it Pharmaceuticals\/} {\bf 12}, 82
  (2019\color{black}).

\bibitem{maccari2013antimicrobial}
G.~Maccari, {\it et~al.\/}, {\it PLoS computational biology\/} {\bf 9},
  e1003212 (2013).

\bibitem{meher2017predicting}
P.~K. Meher, T.~K. Sahu, V.~Saini, A.~R. Rao, {\it Scientific reports\/} {\bf
  7}, 42362 (2017\color{black}).

\bibitem{thomas2010camp}
S.~Thomas, S.~Karnik, R.~S. Barai, V.~K. Jayaraman, S.~Idicula-Thomas, {\it
  Nucleic acids research\/} {\bf 38}, D774 (2010\color{black}).

\bibitem{witten2019deep}
J.~Witten, Z.~Witten, {\it bioRxiv\/}  (2019\color{black}).

\bibitem{xiao2013iamp}
X.~Xiao, P.~Wang, W.-Z. Lin, J.-H. Jia, K.-C. Chou, {\it Analytical
  biochemistry\/} {\bf 436}, 168 (\color{black}2013).

\bibitem{veltri2018deep}
D.~Veltri, U.~Kamath, A.~Shehu, {\it Bioinformatics\/} {\bf 34}, 2740
  (2018\color{black}).

\bibitem{porto2018silico}
W.~F. Porto, {\it et~al.\/}, {\it Nature communications\/} {\bf 9}, 1490
  (2018\color{black}).

\bibitem{fjell2011optimization}
C.~D. Fjell, H.~Jenssen, W.~A. Cheung, R.~E. Hancock, A.~Cherkasov, {\it
  Chemical biology \& drug design\/} {\bf 77}, 48 (2011\color{black}).

\bibitem{porto2018joker}
W.~F. Porto, I.~C. Fensterseifer, S.~M. Ribeiro, O.~L. Franco, {\it Biochimica
  et Biophysica Acta (BBA)-General Subjects\/}  (2018).

\bibitem{nagarajan2019omega76}
D.~Nagarajan, {\it et~al.\/}, {\it Science advances\/} {\bf 5}, eaax1946
  (2019\color{black}).

\bibitem{muller2018recurrent}
A.~T. Mueller, J.~A. Hiss, G.~Schneider, {\it Journal of Chemical Information
  and Modeling\/}  (2018).

\bibitem{grisoni2018designing}
F.~Grisoni, {\it et~al.\/}, {\it ChemMedChem\/} {\bf 13}, 1300
  (2018\color{black}).

\bibitem{gupta2018generative}
A.~Gupta, {\it et~al.\/}, {\it Molecular informatics\/} {\bf 37}, 1700111
  (2018).

\bibitem{gomez1610automatic}
R.~G{\'o}mez-Bombarelli, {\it et~al.\/}, {\it ACS central science\/} {\bf 4},
  268 (2018).

\bibitem{jin2018junction}
W.~Jin, R.~Barzilay, T.~Jaakkola, {\it International Conference on Machine
  Learning\/} (2018), pp. 2323--2332.

\bibitem{blaschke2018application}
T.~Blaschke, M.~Olivecrona, O.~Engkvist, J.~Bajorath, H.~Chen, {\it Molecular
  informatics\/} {\bf 37}, 1700123 (2018).

\bibitem{chan2019advancing}
H.~S. Chan, H.~Shan, T.~Dahoun, H.~Vogel, S.~Yuan, {\it Trends in
  pharmacological sciences\/} {\bf 40}, 592 (2019).

\bibitem{sanchez2018inverse}
B.~Sanchez-Lengeling, A.~Aspuru-Guzik, {\it Science\/} {\bf 361}, 360 (2018).

\bibitem{nagarajan2018computational}
D.~Nagarajan, {\it et~al.\/}, {\it Journal of Biological Chemistry\/} {\bf
  293}, 3492 (2018).

\bibitem{hinton2006reducing}
G.~E. Hinton, R.~R. Salakhutdinov, {\it Science\/} {\bf 313}, 504 (2006).

\bibitem{kingma2013auto}
D.~P. Kingma, M.~Welling, {\it Stat\/} {\bf 1050}, 1 (2014).

\bibitem{guimaraes2017objective}
G.~L. Guimaraes, B.~Sanchez-Lengeling, C.~Outeiral, P.~L.~C. Farias,
  A.~Aspuru-Guzik, {\it arXiv preprint arXiv:1705.10843\/}  (2017).

\bibitem{popova2018deep}
M.~Popova, O.~Isayev, A.~Tropsha, {\it Science advances\/} {\bf 4}, eaap7885
  (2018).

\bibitem{kang2018conditional}
S.~Kang, K.~Cho, {\it Journal of chemical information and modeling\/} {\bf 59},
  43 (2018).

\bibitem{Losasso2019}
V.~Losasso, Y.-W. Hsiao, F.~Martelli, M.~D. Winn, J.~Crain, {\it Physical
  Review Letters\/} {\bf 122}, 208103 (2019).

\bibitem{Cipcigan2018}
F.~Cipcigan, {\it et~al.\/}, {\it The Journal of Chemical Physics\/} {\bf 148},
  241744 (2018).

\bibitem{uniprot}
P.~EMBL-EBI, SIB, {Universal Protein Resource (UniProt)},
  \url{https://www.uniprot.org} (2018). [Online; accessed August-2018].

\bibitem{ElmoPeters:2018}
M.~E. Peters, {\it et~al.\/}, {\it Proc. of NAACL\/} (2018).

\bibitem{gpt2:radford2019language}
A.~Radford, {\it et~al.\/}, {\it OpenAI Blog\/} {\bf 1} (2019).

\bibitem{cove:mccann2017learned}
B.~McCann, J.~Bradbury, C.~Xiong, R.~Socher, {\it Advances in Neural
  Information Processing Systems\/} (2017), pp. 6297--6308.

\bibitem{devlin2018bert}
J.~Devlin, M.-W. Chang, K.~Lee, K.~Toutanova, {\it arXiv preprint
  arXiv:1508.05326\/}  (2018).

\bibitem{rao2019evaluating}
R.~Rao, {\it et~al.\/}, {\it arXiv preprint arXiv:1906.08230\/}  (2019).

\bibitem{Madani2020.03.07.982272}
A.~Madani, {\it et~al.\/}, {\it arXiv preprint arXiv:2004.03497\/}  (2020).

\bibitem{riesselman2017deep}
A.~J. Riesselman, J.~B. Ingraham, D.~S. Marks, {\it Nature methods\/} {\bf 15},
  816 (2018).

\bibitem{shi2016does}
X.~Shi, I.~Padhi, K.~Knight, {\it Proceedings of the 2016 Conference on
  Empirical Methods in Natural Language Processing\/} (2016), pp. 1526--1534.

\bibitem{yu2003compositional}
Y.-K. Yu, J.~C. Wootton, S.~F. Altschul, {\it Proceedings of the National
  Academy of Sciences\/} {\bf 100}, 15688 (2003\color{black}).

\bibitem{vishnepolsky2018predictive}
B.~Vishnepolsky, {\it et~al.\/}, {\it Journal of chemical information and
  modeling\/} {\bf 58}, 1141 (2018\color{black}).

\bibitem{gupta2013silico}
S.~Gupta, {\it et~al.\/}, {\it PloS one\/} {\bf 8}, e73957 (2013).

\bibitem{gomez2018automatic}
R.~G{\'o}mez-Bombarelli, {\it et~al.\/}, {\it ACS central science\/} {\bf 4},
  268 (2018).

\bibitem{sattarov2019novo}
B.~Sattarov, {\it et~al.\/}, {\it Journal of chemical information and
  modeling\/} {\bf 59}, 1182 (2019\color{black}).

\bibitem{pearson2013introduction}
W.~R. Pearson, {\it Current protocols in bioinformatics\/} {\bf 42}, 3 (2013).

\bibitem{li2016molecular}
R.-F. Li, {\it et~al.\/}, {\it Interdisciplinary Sciences: Computational Life
  Sciences\/} {\bf 8}, 319 (2016).

\bibitem{hawrani2008origin}
A.~Hawrani, R.~A. Howe, T.~R. Walsh, C.~E. Dempsey, {\it Journal of Biological
  Chemistry\/} {\bf 283}, 18636 (2008\color{black}).

\bibitem{wiradharma2013rationally}
N.~Wiradharma, M.~Y. Sng, M.~Khan, Z.-Y. Ong, Y.-Y. Yang, {\it Macromolecular
  rapid communications\/} {\bf 34}, 74 (2013\color{black}).

\bibitem{rifkind1967prevention}
D.~Rifkind, {\it Journal of Bacteriology\/} {\bf 93}, 1463 (1967).

\bibitem{ronvcevic2018parallel}
T.~Ron{\v{c}}evi{\'c}, {\it et~al.\/}, {\it BMC genomics\/} {\bf 19}, 827
  (2018).

\bibitem{jing2003conformation}
W.~Jing, A.~R. Demcoe, H.~J. Vogel, {\it Journal of bacteriology\/} {\bf 185},
  4938 (2003).

\bibitem{haney2013mechanism}
E.~F. Haney, {\it et~al.\/}, {\it Biochimica et Biophysica Acta
  (BBA)-Biomembranes\/} {\bf 1828}, 1802 (2013).

\bibitem{mathur2018silico}
D.~Mathur, S.~Singh, A.~Mehta, P.~Agrawal, G.~P. Raghava, {\it PloS one\/} {\bf
  13}, e0196829 (2018).

\bibitem{kumar2018antimicrobial}
P.~Kumar, J.~N. Kizhakkedathu, S.~K. Straus, {\it Biomolecules\/} {\bf 8}, 4
  (2018\color{black}).

\bibitem{guha2019mechanistic}
S.~Guha, J.~Ghimire, E.~Wu, W.~C. Wimley, {\it Chemical reviews\/} {\bf 119},
  6040 (2019).

\bibitem{stokes2020deep}
J.~M. Stokes, {\it et~al.\/}, {\it Cell\/} {\bf 180}, 688 (2020).

\bibitem{loose2006linguistic}
C.~Loose, K.~Jensen, I.~Rigoutsos, G.~Stephanopoulos, {\it Nature\/} {\bf 443},
  867 (2006).

\bibitem{chenthamarakshan2020target}
V.~Chenthamarakshan, {\it et~al.\/}, {\it arXiv preprint arXiv:2004.01215\/}
  (2020).

\bibitem{bowman2015large}
S.~R. Bowman, G.~Angeli, C.~Potts, C.~D. Manning, {\it arXiv preprint
  arXiv:1508.05326\/}  (2015).

\bibitem{hochreiter1997long}
S.~Hochreiter, J.~Schmidhuber, {\it Neural computation\/} {\bf 9}, 1735 (1997).

\bibitem{bowman2015generating}
S.~Bowman, {\it et~al.\/}, {\it Proceedings of The 20th SIGNLL Conference on
  Computational Natural Language Learning\/} (2016), pp. 10--21.

\bibitem{tolstikhin2017wasserstein}
I.~Tolstikhin, O.~Bousquet, S.~Gelly, B.~Sch{\"o}lkopf, {\it Stat\/} {\bf
  1050}, 12 (2018).

\bibitem{bahuleyan2018probabilistic}
H.~Bahuleyan, L.~Mou, K.~Vamaraju, H.~Zhou, O.~Vechtomova, {\it arXiv preprint
  arXiv:1806.08462\/}  (2018).

\bibitem{makhzani2015adversarial}
A.~Makhzani, J.~Shlens, N.~Jaitly, I.~Goodfellow, B.~Frey, {\it arXiv preprint
  arXiv:1511.05644\/}  (2015).

\bibitem{gretton2007kernel}
A.~Gretton, K.~M. Borgwardt, M.~Rasch, B.~Sch{\"o}lkopf, A.~J. Smola, {\it
  Advances in neural information processing systems\/} (2007), pp. 513--520.

\bibitem{rahimi2007random}
A.~Rahimi, B.~Recht, {\it Advances in neural information processing systems\/}
  (2007), pp. 1177--1184.

\bibitem{rubenstein2018latent}
P.~K. Rubenstein, B.~Schoelkopf, I.~Tolstikhin, {\it arXiv preprint
  arXiv:1802.03761\/}  (2018).

\bibitem{yu2017seqgan}
L.~Yu, W.~Zhang, J.~Wang, Y.~Yu, {\it Thirty-First AAAI Conference on
  Artificial Intelligence\/} (2017).

\bibitem{jang2016categorical}
E.~Jang, S.~Gu, B.~Poole, {\it arXiv preprint arXiv:1611.01144\/}  (2016).

\bibitem{kusner2016gans}
M.~J. Kusner, J.~M. Hern{\'a}ndez-Lobato, {\it arXiv preprint
  arXiv:1611.04051\/}  (2016).

\bibitem{maddison2016concrete}
C.~J. Maddison, A.~Mnih, Y.~W. Teh, {\it arXiv preprint arXiv:1611.00712\/}
  (2016).

\bibitem{zhang2016generating}
Y.~Zhang, Z.~Gan, L.~Carin, {\it NIPS workshop on Adversarial Training\/}
  (2016).

\bibitem{kingma2014semi}
D.~P. Kingma, S.~Mohamed, D.~J. Rezende, M.~Welling, {\it Advances in Neural
  Information Processing Systems\/} (2014), pp. 3581--3589.

\bibitem{hu2017toward}
Z.~Hu, Z.~Yang, X.~Liang, R.~Salakhutdinov, E.~P. Xing, {\it International
  Conference on Machine Learning\/} (2017), pp. 1587--1596.

\bibitem{engel2017latent}
J.~Engel, M.~Hoffman, A.~Roberts, {\it arXiv preprint arXiv:1711.05772\/}
  (2017).

\bibitem{dathathri2019plug}
S.~Dathathri, {\it et~al.\/}, {\it arXiv preprint arXiv:1912.02164\/}  (2019).

\bibitem{zhou2019optimization}
Z.~Zhou, S.~Kearnes, L.~Li, R.~N. Zare, P.~Riley, {\it Scientific reports\/}
  {\bf 9}, 10752 (2019).

\bibitem{you2018graph}
J.~You, B.~Liu, Z.~Ying, V.~Pande, J.~Leskovec, {\it Advances in Neural
  Information Processing Systems\/} (2018), pp. 6410--6421.

\bibitem{Zhavoronkov2019natbio}
A.~Zhavoronkov, {\it et~al.\/}, {\it Nature Biotechnology\/} {\bf 37}, 1038
  (2019).

\bibitem{korovina2019chembo}
K.~Korovina, {\it et~al.\/}, {\it arXiv preprint arXiv:1908.01425\/}  (2019).

\bibitem{lim2018molecular}
J.~Lim, S.~Ryu, J.~W. Kim, W.~Y. Kim, {\it Journal of cheminformatics\/} {\bf
  10}, 31 (2018).

\bibitem{li2018multi}
Y.~Li, L.~Zhang, Z.~Liu, {\it Journal of cheminformatics\/} {\bf 10}, 33
  (2018).

\bibitem{mendez2020novo}
O.~M{\'e}ndez-Lucio, B.~Baillif, D.-A. Clevert, D.~Rouqui{\'e}, J.~Wichard,
  {\it Nature Communications\/} {\bf 11}, 1 (2020).

\bibitem{singh2015satpdb}
S.~Singh, {\it et~al.\/}, {\it Nucleic acids research\/}  (2015).

\bibitem{pirtskhalava2016dbaasp}
M.~Pirtskhalava, {\it et~al.\/}, {\it Nucleic acids research\/} {\bf 44}, D1104
  (2016).

\bibitem{khurana2018deepsol}
S.~Khurana, {\it et~al.\/}, {\it Bioinformatics\/} {\bf 34}, 2605 (2018).

\bibitem{bhadra2018ampep}
P.~Bhadra, J.~Yan, J.~Li, S.~Fong, S.~W. Siu, {\it Scientific reports\/}
  (2018).

\bibitem{Frishman_Argos_1995}
D.~Frishman, P.~Argos, {\it Proteins\/} {\bf 23}, 566–579 (1995).

\bibitem{theis2015note}
L.~Theis, A.~v.~d. Oord, M.~Bethge, {\it ICLR\/}  (2016).

\bibitem{alemi2017fixing}
A.~A. Alemi, {\it et~al.\/}, {\it arXiv preprint arXiv:1711.00464\/}  (2017).

\bibitem{sercu2019interactive}
T.~Sercu, {\it et~al.\/}, {\it ICLR Workshop on Deep Generative Models for
  Highly Structured Data\/}  (2019).

\bibitem{ranzato2015sequence}
M.~Ranzato, S.~Chopra, M.~Auli, W.~Zaremba, {\it arXiv preprint
  arXiv:1511.06732\/}  (2015).

\bibitem{bengio2015scheduled}
S.~Bengio, O.~Vinyals, N.~Jaitly, N.~Shazeer, {\it Advances in Neural
  Information Processing Systems\/} (2015), pp. 1171--1179.

\bibitem{zhao2017adversarially}
J.~J. Zhao, Y.~Kim, K.~Zhang, A.~Rush, Y.~LeCun, {\it arXiv preprint
  arXiv:1706.04223\/}  (2017).

\bibitem{meritylm}
S.~Merity, N.~S. Keskar, R.~Socher, {\it arXiv preprint arXiv:1708.02182\/}
  (2017).

\bibitem{Tien2013}
M.~Z. Tien, D.~K. Sydykova, A.~G. Meyer, C.~O. Wilke, {\it {PeerJ}\/} {\bf 1},
  e80 (2013).

\bibitem{deJong2012}
D.~H. de~Jong, {\it et~al.\/}, {\it Journal of Chemical Theory and
  Computation\/} {\bf 9}, 687 (2012).

\bibitem{Wassenaar2015}
T.~A. Wassenaar, H.~I. Ing{\'{o}}lfsson, R.~A. B\"{o}ckmann, D.~P. Tieleman,
  S.~J. Marrink, {\it Journal of Chemical Theory and Computation\/} {\bf 11},
  2144 (2015).

\bibitem{Marrink2007}
S.~J. Marrink, H.~J. Risselada, S.~Yefimov, D.~P. Tieleman, A.~H. de~Vries,
  {\it The Journal of Physical Chemistry B\/} {\bf 111}, 7812 (2007).

\bibitem{Berendsen1995}
H.~Berendsen, D.~van~der Spoel, R.~van Drunen, {\it Computer Physics
  Communications\/} {\bf 91}, 43 (1995).

\bibitem{Abraham2015}
M.~J. Abraham, {\it et~al.\/}, {\it {SoftwareX}\/} {\bf 1-2}, 19 (2015).

\bibitem{Bussi2007}
G.~Bussi, D.~Donadio, M.~Parrinello, {\it The Journal of Chemical Physics\/}
  {\bf 126}, 014101 (2007).

\bibitem{Parrinello1981}
M.~Parrinello, A.~Rahman, {\it Journal of Applied Physics\/} {\bf 52}, 7182
  (1981).

\bibitem{Nos1983}
S.~Nos{\'{e}}, M.~Klein, {\it Molecular Physics\/} {\bf 50}, 1055 (1983).

\bibitem{Huang2016}
J.~Huang, {\it et~al.\/}, {\it Nature Methods\/} {\bf 14}, 71 (2016).

\bibitem{qin2020artificial}
Z.~Qin, {\it et~al.\/}, {\it Extreme Mechanics Letters\/} p. 100652 (2020).

\bibitem{Jo2009}
S.~Jo, J.~B. Lim, J.~B. Klauda, W.~Im, {\it Biophysical Journal\/} {\bf 97}, 50
  (2009).

\bibitem{HUMP96}
W.~Humphrey, A.~Dalke, K.~Schulten, {\it Journal of Molecular Graphics\/} {\bf
  14}, 33 (1996).

\bibitem{Phillips2005}
J.~C. Phillips, {\it et~al.\/}, {\it Journal of Computational Chemistry\/} {\bf
  26}, 1781 (2005).

\bibitem{muller2017modlamp}
A.~T. M{\"u}ller, G.~Gabernet, J.~A. Hiss, G.~Schneider, {\it Bioinformatics\/}
  {\bf 33}, 2753 (2017).

\bibitem{cock2009biopython}
P.~J. Cock, {\it et~al.\/}, {\it Bioinformatics\/} {\bf 25}, 1422
  (2009\color{black}).

\bibitem{madden2013blast}
T.~Madden, {\it The NCBI Handbook [Internet]. 2nd edition\/} (National Center
  for Biotechnology Information (US), 2013\color{black}).

\bibitem{altschul1990basic}
S.~F. Altschul, W.~Gish, W.~Miller, E.~W. Myers, D.~J. Lipman, {\it Journal of
  molecular biology\/} {\bf 215}, 403 (1990\color{black}).

\bibitem{Patseq}
{PATSEQ}, \url{https://www.lens.org/lens/bio/}. [Online].

\bibitem{chin2018macromolecular}
W.~Chin, {\it et~al.\/}, {\it Nature communications\/} {\bf 9}, 1 (2018).

\bibitem{ng2013synergistic}
V.~W.~L. Ng, X.~Ke, A.~L. Lee, J.~L. Hedrick, Y.~Y. Yang, {\it Advanced
  Materials\/} {\bf 25}, 6730 (2013).

\bibitem{liu2017highly}
S.~Liu, {\it et~al.\/}, {\it Biomaterials\/} {\bf 127}, 36 (2017).

\end{thebibliography}

\clearpage

\section*{Table of Contents}

Supplementary Text\\
Dataset\\
Model and Methods\\
Fig. S1 to S4 \\
Tables S1 to S8\\
Supplementary References \textit{(1-67)}
\section{Supplementary Text}

\subsection{Conditional Generation with Autoencoders}
Since the peptide sequences are represented as text strings here, we will be limiting our discussion to literature around text generation with constraints. 
Controlled text sequence generation is non-trivial, as the discrete and non-differentiable nature of text samples does not allow the use of a global discriminator, which is commonly used in image generation tasks to guide generation. To tackle this issue of non-differentiability, policy learning has been suggested, which suffers from high variance during training \cite{yu2017seqgan,guimaraes2017objective}. Therefore, specialized distributions, such as the Gumbel-softmax \cite{jang2016categorical,kusner2016gans}, a concrete distribution \cite{maddison2016concrete},  or a soft-argmax function \cite{zhang2016generating}, have been proposed to approximate the gradient of the model from discrete samples.

Alternatively, in a semi-supervised model setting, the minimization of element-wise reconstruction error has been employed \cite{kingma2014semi}, which tends to lose the holistic view of a full sentence. Hu et al. \cite{hu2017toward} proposed a VAE variant that allows both controllable generation and semi-supervised learning. The working principle needs labels to be present during training and encourages latent space to represent them, so the addition of  
 new attributes will require retraining the latent variable model itself.
The framework relies on a set of new discrete binary variables in latent space to control the attributes, an ad-hoc wake-sleep procedure. It requires learning the right balance between multiple competing and tightly interacting loss objectives, which is tricky. 

Engel et al.~\cite{engel2017latent} propose a conditional generation framework without retraining the model, similar in concept to ours, by modeling in latent space post-hoc. Their approach does not need an explicit density model in $\rvz$-space, 
rather relies on adversarial training of generator and attribute discriminator and focuses modifying sample reconstructions rather than generating novel samples. 
Recent Plug and Play Language Model (PPLM) for controllable language generation combines a pre-trained language model (LM) with one or more simple attribute classifiers that guide text generation without any further training of the LM \cite{dathathri2019plug}. 

\subsection{ Conditional Generation for Molecule Design}
Following the work of Gómez-Bombarelli et al. \cite{gomez2018automatic}, Bayesian Optimization (BO) in the learned latent space has been employed for molecular optimization for properties such as drug likeliness (QED) or penalized logP. The standard BO routine consists of two key steps: (i) estimating the black-box function from data through a probabilistic surrogate model; usually a Gaussian process (GP), referred to as the response surface; (ii) maximizing an acquisition function that computes a score that trades off exploration and exploitation according to uncertainty and optimality of the response surface. As the dimensionality of the input latent space increases, these two steps become challenging.  In most cases, such a method is restricted to local optimization using training data points as starting points, as optimizers are likely to follow gradients into regions of the latent space that the model has not been exposed to during training.
Reinforcement learning (RL) based methods provide an alternative approach for molecular optimization \cite{zhou2019optimization, you2018graph, popova2018deep, Zhavoronkov2019natbio}, in which RL policies are learned by incorporating the desired attribute as part of the reward. However, a large number of evaluations are typically needed for both BO and RL-based optimizations while trading off exploration and exploitation \cite{korovina2019chembo}. 
Semi-supervised learning has also been used for conditional generation of molecules \cite{lim2018molecular, kang2018conditional, li2018multi, mendez2020novo}, which needs labels to be available during the generative model training.  

CLaSS is fundamentally different from these existing approaches, as it does not need expensive optimization over latent space, policy learning, or minimization of complex loss objectives - and therefore does not suffer from cumbersome computational complexity. Furthermore, CLaSS is not limited to local optimization around an initial starting point. Adding a new constraint in CLaSS is relatively simple, as it only requires a simple predictor training; therefore, CLaSS is easily repurposable. CLaSS is embarrassingly parallelizable as well.

\section{Dataset}

\subsection{A Dataset for Semi-Supervised Training of AMP Generative Model}
\label{si:data}

We compiled a new two-part (unlabeled and labeled) dataset for learning a meaningful representation of the peptide space and conditionally generating safe antimicrobial peptides from that space using the proposed CLaSS method. We consider discriminating for several functional attributes as well as for presence of structure in peptides. Only linear and monomeric sequences with no terminal modifications and length up to 50 amino acids were considered in curating this dataset.
As a further pre-processing step, the sequences with non-natural amino acids (B, J, O, U, X, and Z) and the ones with lower case letters were eliminated. 

\textbf{Unlabeled Sequences:} The unlabeled data is from Uniprot-SwissProt and Uniprot-Trembl database \cite{uniprot} and contains just over 1.7 M  sequences, when considering sequences with length up to 50 amino acid.

\textbf{Labeled Sequences:}
Our labeled dataset comprises sequences with different attributes curated from a number of publicly available databases~\cite{singh2015satpdb,pirtskhalava2016dbaasp, khurana2018deepsol, bhadra2018ampep, gupta2013silico}. Below we provide details of the labeled dataset:
\begin{itemize}
    \item Antimicrobial (8683 AMP, 6536 non-AMP);
    \item Toxic (3149 Toxic, 16280 non-Toxic);
    \item Broad-spectrum (1302 Positive, 1238 Negative);
    \item Structured (1170 Positive, 2136 Negative);
    \item Hormone (569 Positive);
    \item Antihypertensive (1659 Positive);
    \item Anticancer (504 Positive).
\end{itemize}

\subsubsection{Details of Labeled Datasets}
\label{si:datadetails}
\textbf{Sequences with AMP/non-AMP Annotation.}
AMP labeled dataset comprises sequences
from two major AMP databases:  satPDB \cite{singh2015satpdb} and DBAASP \cite{pirtskhalava2016dbaasp}, as well as a dataset used in an earlier AMP classification study named as AMPEP \cite{bhadra2018ampep}. 
Sequences with an antimicrobial function annotation in satPDB and AMPEP or a MIC value against any target species less than 25 $\mu$g$/$ml in DBAASP 
were considered as   AMP labeled instances. The duplicates between these three datasets were removed to generate a non-redundant AMP dataset. And, the ones with mean activity against all target species  $>$ 100 $\mu$g$/$ml in DBAASP were considered negative instances (non-AMP). Since experimentally verified non-AMP sequences are rare to find, the non-AMP instances in AMPEP were generated from UniProt sequences after discarding sequences that were annotated as AMP, membrane, toxic, secretory, defensive, antibiotic, anticancer, antiviral, and antifungal and were used in this study as well.


\textbf{Sequences with Toxic/nonToxic Annotation:}
Sequences with toxicity labels are curated from satPDB and DBAASP databases as well as from the ToxinPred dataset \cite{gupta2013silico}. Sequences with ``Major Function" or ``Sub Function" annotated as toxic in satPDB and sequences with hemolytic/cytotoxic activities against all reported target species less than 200 $\mu$g$/$ml in DBAASP were considered as Toxic instances. The toxic-annotated instances from ToxinPred were added to this set after removing duplicates resulting in a total to 3149 Toxic sequences. 
Sequences  with hemolytic/cytotoxic activities $>$ 250 $\mu$g$/$ml were considered as nonToxic. The nonToxic instances reported in ToxinPred (sequences from SwissProt or TrEMBL that are not found in search using keyword associated with toxins, \textit{i.e.} keyword (NOT KW800 NOT KW20) or keyword (NOT KW800 AND KW33090), were added to the nonToxic set, totaling to 16280 non-AMP sequences.  

\textbf{Sequences with Broad-Spectrum  Annotation}
Antimicrobial sequences reported in the satPDB or DBAASP database can have both Gram-positive and Gram-negative strains as target groups. We consider such sequences as broad-spectrum. Otherwise, they are treated as narrow-spectrum. Through our filtering, we found  1302 broad-spectrum and 1238 narrow-spectrum sequences. 

\textbf{Sequences with Structure/No-Structure Annotation:}
Secondary Structure assignment was performed for structures from satPDB using the STRIDE algorithm~\cite{Frishman_Argos_1995}. If more than 60\% of the amino acids are helix or beta-strand, we label it as structured (positive). Otherwise, they are labeled as negative. Through this filtering, we found 1170 positive sequences and 2136 negative sequences.

\textbf{Peptide Dataset for Baseline Simulations:} For the control simulations, three datasets were prepared. The first two were taken from the satpdb dataset, filtering sequences with length smaller than 20 amino acids. The high potency dataset contains the 51 sequences with the lowest average MIC, excluding sequences with cysteine residues. All these sequences have an average MIC of less than 10 $\mu$g / ml. The low potency dataset contains the 41 sequences with the highest MIC, excluding cysteine residues. All these have an average MIC over 300 $\mu$g / ml.

To create a dataset of inactive sequences, we queried UniProt using the following keywords: \emph{NOT keyword:"Antimicrobial [KW-0929]" length:[1 TO 20] NOT keyword:"Toxin [KW-0800]" NOT keyword:"Disulfide bond [KW-1015]" NOT annotation:(type:ptm) NOT keyword:"Lipid-binding [KW-0446]" NOT keyword:"Membrane [KW-0472]" NOT keyword:"Cytolysis [KW-0204]" NOT keyword:"Cell wall biogenesis/degradation [KW-0961]" NOT keyword:"Amphibian defense peptide [KW-0878]" NOT keyword:"Secreted [KW-0964]" NOT keyword:"Defensin [KW-0211]" NOT keyword:"Antiviral protein [KW-0930]" AND reviewed:yes}. From this, we picked 54 random sequences to act as the inactive dataset for simulation.

\section{Model and Methods}
\subsection{Autoencoder Details}
\label{sec:model}


\subsubsection{Autoencoder Architecture}
\label{si:models}
We investigate two different types of autoencoding approaches: $\beta$-VAE\cite{bowman2015generating} and  WAE\cite{tolstikhin2017wasserstein} in this study. 
For each of these AEs the default architecture involves bidirectional-GRU encoder and GRU decoder. 
For the encoder, we used a bi-directional GRU with hidden state size of 80. 
The latent capacity was set at $D=100$.

For VAE, we used KL term annealing $\beta$ from 0 to 0.03 by default.
We also present an unmodified VAE with KL term annealing $\beta$ from 0 to 1.0. 
For WAE, we found that the random features approximation of the gaussian kernel with kernel bandwidth $\sigma=7$  to be performing the best. For comparison sake, we have included variations with $\sigma$ values of 3 and 15 too. 
The inclusion of $z$-space noise logvar regularization, $R(logVar)$, helped avoiding collapse to a deterministic encoder.
Among different regularization weights used,  $1e-3$ had the most desirable behavior on the metrics (see Section \ref{si:vaewaeaae}).

\subsubsection{Autoencoder Training}
\label{method:training}

When training the AE model, we  sub selected sequences with length $\leq$ the hyperparameter
$max\_seq\_length$.
Furthermore, both AMP-labeled and unlabeled data were split into train, held out, and test set.
This reduces the available sequences for training; \textit{e.g} for unlabeled set the number of available training sequences are 93k for $max\_seq\_length$=25, whereas the number of AMP-labeled sequences was 5000. 
The sequences with reported activities were considered as confirmed labeled data, and those with confirmed labels were up-sampled at a 1:20 ratio. 
Such upsampling of peptides with a specific attribute label will allow mitigation of possible domain shift due to unlabeled peptides coming from a different distribution. 
 However, the
 benefit of transfer learning from unlabeled to labeled data likely outweighs the effects of domain shift.

To obtain the optimal hyperparameter setting for autoencoder training, we adopted an automated hyperparameter optimization.  Specifically, we performed a grid search in the hyperparameter space and tracked an $L_2$ distance between the reconstructions of held-out data and training sequences, which was estimated using a weighted combination of BLEU, PPL, $\rvz$-classifier accuracy, and amino acid composition-based heuristics. 
The best hyperparameter configuration obtained using this process was the following: learning rate = 0.001, number of iterations = 200000, minibatch size = 32, word dropout = 0.3. A beam search decoder was used with a beam size of 5. 

\subsubsection{ Autoencoder Evaluation}
Evaluation of generative models is notoriously difficult \cite{theis2015note}.
In the variational auto-encoder family, two competing objective terms are minimized: reconstruction of the input and a form of regularization in the latent space, which form a fundamental trade-off \cite{alemi2017fixing}.
Since we want a meaningful and consistent latent space,  models that do not compromise the reconstruction quality to achieve lower constraint loss are preferred.
We propose an evaluation protocol using four metrics to judge the quality of both heldout reconstructions and prior samples \cite{sercu2019interactive}. The metrics are
\begin{enumerate}[(i)]
    \item The objective terms, evaluated on heldout data: reconstruction log likelihood $-\log p_\theta(\rvx|\rvz)$ and $D(q^h_\phi(\rvz) | p(\rvz))$, where $q^h_\phi(\rvz) = \frac{1}{N_{hld}}\sum_{\rvx^i \sim \text{hld}} q_\phi(\rvz|\rvx^i)$ is the average over heldout encodings.
    \item Encoder variance $\log(\sigma_j^2(\rvx^i))$ averaged over heldout samples, in $\normltwo$ over components $j$. In order to achieve a meaningful latent space, we needed to regularize the encoder variance to not becoming vanishingly small, i.e., for the encoder to become deterministic \cite{rubenstein2018latent}. Large negative values indicate that the encoder is collapsed to deterministic.
    \item Reconstruction BLEU score on held-out samples.
    \item Perplexity (PPL) evaluated by an external language model, for samples from prior $p(\rvz)$ and heldout encoding $q_\phi(\rvz|\rvx)$.
\end{enumerate}
Note that (iii) and (iv) involve sampling the decoder (we use beam search with beam 5), which will therefore also take into account any exposure bias \cite{ranzato2015sequence,bengio2015scheduled}. 
We propose to evaluate peptide generation using the perplexity under an independently trained language model (iv), which is a reasonable heuristic \cite{zhao2017adversarially} if we assume the independent language model captures the distribution $p(\rvx)$ well. Perplexity of a language model is the inverse probability of the test set, normalised by the number of words. 
A lower perplexity indicates a better model.

\textbf{External Language Model}

We trained an external language model on both labeled and unlabeled sequences to determine the perplexity of the generated sequences. Specifically, we used a character-based LSTM language model (LM) with the help of LSTM and QRNN Language Model Toolkit\cite{meritylm} trained on both AMP-labeled and unlabeled data. We trained our language model with a total of $92624$ sequences, with a maximum sequence length of $25$. Our best model achieves a test perplexity of $13.26$. 

To further validate the performance of our language model, we tested it on sequences with randomly generated synthetic amino acids from the vocabulary of lengths ranging between $10$ to $25$. As expected, we found it to have a high perplexity of $27.29$. Also, when evaluating it for repeated amino acids (sequence consisting of a single token from vocabulary), of length ranging between 10 to 25, we found the perplexity to be very low ($3.48$). Upon further investigation, we observed that the training data consists of amino acids with repeated sub-sequences, which the language model by nature, fails to penalize heavily. Due to this behavior, we can conclude that the perplexity of a collapsed peptide model will be closer to $3.48$ (as seen in the case of $\beta$-VAE). We have summarized these observation in Table \ref{tab:model_results}.

\subsubsection{Autoencoder Variants: Comparison}
\label{si:vaewaeaae}
\begin{table}[ht]
\centering
\begin{tabular}{l|l|rrrr}
\toprule
\multicolumn{2}{c}{Architecture} & PPL & BLEU & Recon & Encoder variance  \\ \hline
\multirow{4}{*}{Reference}   & Repeated Sequences   & 3.48 &  & &    \\   
                      &  Random  & 27.29  &   &  & \\   
                      &  AMP Labeled  & 5.580  &   &  &  \\   
                      &  Labeled + Unlabeled  & 13.26  &  &  & \\
                      &  Peptide LSTM,  \cite{muller2018recurrent} (Labeled)  & 22.40  &  &  & \\
                      &  Peptide LSTM,  \cite{muller2018recurrent} (Unlabeled)  & 20.26  &  &  & \\\hline
 \multicolumn{2}{l}{$\beta$-VAE (1.0)}                   &	3.820   &   4e-03   & 2.768 & -2.5e-4   \\ 
\multicolumn{2}{l}{$\beta$-VAE (0.03)}                  &	15.34   &   0.475   & 1.075 & -0.620  \\
\cmidrule{1-6} \multicolumn{2}{l}{WAE, $\sigma=3$, $R(logVar)=1e-3$}           &	13.25 & 0.853  & 0.257 & -3.078 \\
\multicolumn{2}{l}{WAE, $\sigma=15$, $R(logVar)=1e-3$}                         &	12.98 & 0.909  & 0.224 & -4.180 \\
\multicolumn{2}{l}{WAE, $\sigma=7$, $R(logVar) =0$}         &	12.77 & 0.881  & 0.214 & -13.81 \\
\multicolumn{2}{l}{WAE, $\sigma=7$, $R(logVar)=1e-2$}      &	15.16 & 0.665  & 0.685 & -0.3962 \\
 \multicolumn{2}{l}{WAE, $\sigma=7$, $R(logVar)=1e-3$}          & 12.87 & 0.892  & 0.216 &  -4.141  \\ 
 \cmidrule{1-6} \multicolumn{2}{l}{WAE (trained on AMP-labeled)} &16.12 & 0.510 &  0.354 & -4.316 \\ 
 \bottomrule
\end{tabular}
\caption{Performance of various autoencoder schemes against different baselines. }
\label{tab:model_results}
\end{table}

The evaluated metrics on held-out samples for different autoencoder models trained on either labeled or full dataset are also reported in Table \ref{tab:model_results}. We observed that the reconstruction of WAE is more accurate compared to $\beta$-VAE: we achieve a  reconstruction error of $0.2163$ and a BLEU score of $0.892$ on a held-out set using WAE with $\sigma$ of 7 and $R(logVar)$ of 1e-3 (values are $1.079$ and $0.493$ for $\beta$-VAE with $\beta$ set to $0.03$). The advantage of using abundant unlabeled data compared to only the labeled ones for representation learning is evident,  as the language model perplexity (PPL, captures sequence diversity) for the WAE model trained on a full dataset is closer to that of the test perplexity ($13.26$), and the BLEU score is also higher when compared to the WAE model trained only on AMP-labeled sequences. For reference, PPL of random peptide sequences is $> 25$, and for repeated sequences is $3.48$. We have also compared the perplexity of the generated sequences using a LSTM model used in \cite{ muller2018recurrent} and have reported the perplexity in Table \ref{tab:model_results}.  Results show that the perplexity of the generated samples using the LSTM model is around 22.4, which is close to the random sequence perplexity. In contrast, WAE (as well as VAE with appropriate KL annealing) achieves much lower perplexity that is close to the perplexity of peptide sequences. 
Attribute-controlled sampling using a simple LSTM model is non-trivial.

The $\rvz$-classifier trained on the latent space of the best WAE model achieved a test accuracy of 87.4, 68.9, 77.4, 98.3, 76.3\% for detecting peptides with AMP/non-AMP, toxic/non-toxic, anticancer, antihypertensive, and hormone annotation. 

\subsection{Post-Generation Screening}

\subsubsection{Sequence-Level Attribute Classifiers}
\label{si:xclf}
For post-generation screening, we used four sets of monolithic classifiers that are trained directly on peptide sequences. Each of these binary sequence-level classifiers was aimed at capturing one of the following four properties of peptide sequences, namely, 
\begin{itemize}
    \item AMP/Non-AMP : Is the sequence an AMP or not?
    \item Toxicity/Non-Toxic : Is the sequence toxic or not?
    \item Broad/Narrow  : Does the sequence show antibacterial activity on both Gram+ and Gram- strains or not?
    \item Structure/No-Structure : Does the sequence have secondary structure or not?  
\end{itemize}{}

For each attribute, we trained a bidirectional LSTM-based classifier on the labeled dataset. 
We used a hidden layer size of 100 and a dropout of 0.3. Size of dataset used as well as accuracies are  reported in the Table \ref{tab:classifier_results}.

\begin{table}[ht]
\centering
\begin{tabular}{l|lcc|cc|c}
\toprule
\multicolumn{1}{c|}{\textbf{Attribute}} &  \multicolumn{3}{c|}{\textbf{Data-Split}} & \multicolumn{2}{|c|}{\textbf{Accuracy (\%)}} & \textbf{Screening}  \\ 
\cmidrule{2-5} \cmidrule{5-6}
{} & \textit{Train} & \textit{Valid} & \textit{Test} & \textit{Majority Class} & \textit{Test} &  \textbf{Threshold} \\ \hline
\big\{AMP , Non-AMP\big\} & 6489 & 811 & 812 &  68.9  & 88.0 & 7.944  \\ 
\big\{Toxic , Non-Toxic\big\} & 8153 & 1019 & 1020 &  82.73  & 93.7 & -1.573  \\
\big\{Broad, Narrow\big\} & 2031 & 254 & 255 & 51.37 & 76.0 & -7.323  \\
\big\{Structure , No-Structure\big\}  & 2644 & 331 & 331 & 64.65  & 95.1 & -5.382  \\
 \bottomrule
\end{tabular}
\caption{Performance of classifiers based on different attributes.}
\label{tab:classifier_results}
\end{table}

\begin{table}[ht]
    \centering
    \begin{tabular}{p{4cm}|p{2cm}|p{3.5cm}|p{4cm}}
        \toprule
        \textbf{Classifiers} & Reported Accuracy &  YI12 & FK13  \\ 
        \hline   
        AMP Classifier (ours) &  88.00 & AMP & AMP \\
        AMP Scanner v2\cite{veltri2018deep} & 91.00 & Non-AMP & AMP \\
        iAMP Pred  \cite{meher2017predicting} &  66.80 & AMP & AMP \\
        DBAASP-SP \cite{vishnepolsky2018predictive} & 79.00 & AMP & Non-AMP \\
        iAMP-2L \cite{xiao2013iamp} &  92.23 & Non-AMP & AMP \\
        CAMP-RF \cite{thomas2010camp} & 87.57 & AMP & AMP \\
        Witten E.Coli \cite{witten2019deep} & 94.30 & AMP & Borderline \footnotemark  \\ 
        Witten S.aureus \cite{witten2019deep} & 94.30 & AMP & AMP \\ 
         \bottomrule
    \end{tabular}
    \caption{Reported accuracy and prediction for YI12 and FK13 using different AMP classifiers: AMP Scanner v2 \cite{veltri2018deep}, iAMP-2L \cite{meher2017predicting}, DBAASP Prediction \cite{vishnepolsky2018predictive}, iAMPpred \cite{xiao2013iamp}, CAMP-RF \cite{thomas2010camp}, Witten (CNN-70\%) \cite{witten2019deep}, and our LSTM-based sequence-based classifier.  
}
    \label{tab:AMP-compare}
\end{table}{}

\footnotetext{Witten model predicts logMIC activity values. We followed their criteria for classification: if logMIC is within -1 to 3.5, then the peptide is  active (AMP), if $>3.9$ then inactive (Non-AMP), and between 3.5 to 3.9 is considered borderline. LogMIC values for YI12 (\textit{E. coli})=1.648,  YI12 (\textit{S. aureus}=3.607, FK13 (\textit{E. coli})=1.389,  FK13 (\textit{S. aureus})=1.634.}

Based on the distribution of the scores (classification probabilities/logits), we determined the threshold by considering the $50^{th}$ percentile (median) of the scores ( reported in the last column of Table \ref{tab:classifier_results}). Similarly, we selected a   PPL threshold of $16.04$ that is the $25^{th}$ percentile of the PPL distribution of samples generated from the prior distribution of the best WAE model and also closer to the perplexity of our trained language model on test data.

\subsubsection{CGMD Simulations - Contact Variance as a Metric for Classifying Membrane Binding}
\label{method:sim}
Given a peptide sequence as an input, PeptideBuilder~\cite{Tien2013} is used to prepare a PDB file of the all-atom representation of the peptide. This is prepared either as an alpha helix (with dihedral angles $\phi=-57$,  $\psi=-47$) or as a random coil, with $\phi$ and $\psi$ dihedral angles taking random values between $-50^\circ$ and $50^\circ$. 

This initial structure is then passed as in input to \texttt{martinize.py}~\cite{deJong2012}, which coarse-grains the system. The resulting files are passed into \texttt{insane.py}~\cite{Wassenaar2015} to create the peptide membrane system. The solvent is a 90:10 ratio of water to antifreeze particles, with the membrane being a 3:1 mixture of POPC to POPG. The system is 15 nm x 15 nm x 30 nm, with the membrane perpendicular to the longest direction. Ions are added to neutralize the system.

This is a minimal model of a bacterial membrane that serves as a high-throughput filter. While this model does not replicate the complex physics of the peptide-membrane interactions in exact detail (e.g. membrane composition difference between Gram-positive \textit{vs.} Gram-negative), it allows us to prioritise the peptides that show the strongest interaction with a membrane for experimental characterisation.

For the CGMD simulations, we used the Martini forcefield~\cite{Marrink2007},
as Martini is optimized for predicting the interactions between proteins and membranes while being computationally efficient,  it is well suited for the task of a quick but physically-inspired filtering peptide sequences.

After building, the system is minimized for 50,000 steps using Gromacs 2019.1~\cite{Berendsen1995, Abraham2015} and the 2.0 version of the Martini forcefield \cite{Marrink2007}. After minimization, the production run is carried for 1 $\mu$s at a 20 fs timestep. Temperature is kept constant at 310 K using Stochastic Velocity Rescaling \cite{Bussi2007} applied independently to the protein, lipid, and the solvent groups. The pressure is kept constant at 1 atmosphere using a Parrinello-Rahman barostat~\cite{Parrinello1981, Nos1983} applied independently to the same groups as the thermostat. 

After 1 $\mu$s of sampling, we estimated the number of peptide-membrane contacts using TCL scripting and in-house Python scripts. 
The number of contacts between positive residues and the lipid membranes is defined as the number of atoms belonging to a lipid at a distance less than 7.5 \AA{} from a positive residue. 

For the control simulations, three datasets consisting of reported high-potency AMP, low-potency AMP, and non-AMP sequences were used that are discussed in \ref{si:datadetails}. We performed a set of 130 control simulations. We found that the variance of the number of contacts (cutoff 7.5 \AA{}) between positive residues and Martini beads of the membrane lipids is predictive of antimicrobial activity. Specifically, the contact variance distinguishes between high potency and non-antimicrobial sequences with a sensitivity of 88\% and specificity of 63\%. 
To screen, we used a cutoff value of 2 beads for the contact variance. We carried out a set of simulations for the 163 amp-positive and 179 amp-negative generated sequences. 
We further restricted to sequences that bind in less than $500$ ns during the $1\,\mu$s long simulation, so that the contact variance is calculated over at least half of the total simulation time. Only sequences that formed at least $5$ contacts (averaged over the duration of the simulation) were considered.

\subsection{All-Atom Simulations}
\label{method:aa}

We used the CHARMM36m \cite{Huang2016} forcefield to simulate the binding of four copies of YL12 and FK13 to a model membrane. Phi and Psi angles in the initial peptide structure were set to what was predicted using a recent deep learning model \cite{qin2020artificial}.  A 3:1 DLPC:DLPG bilayer, with shorter tails, was used to speed up the simulation, alongside a smaller water box (98 \AA{}) than the Martini simulations, to investigate the short-term effect of peptide-membrane interactions.  A 160 ns long trajectory was run for the FK13 system. The length of the YI12 simulation was 200 ns. The number of peptide-membrane and peptide-peptide contacts (using a threshold of 7.5 \AA{} and ignoring hydrogen atoms) were found to be converged in less time than the maximum simulation length for both systems. 

The bilayer is prepared using CHARMM-GUI~\cite{Jo2009}, and the peptide sequence is prepared using PeptideBuilder~\cite{Tien2013}. Solvation and assembly is performed using VMD 1.9.3 \cite{HUMP96}. The system is simulated using NAMD 2.13 \cite{Phillips2005}. Temperature is kept constant using a Langevin thermostat and a Nos\'e-Hoover Langevin piston barostat
The Particle-Mesh Ewald method was used for long-range electrostatics. All simulations used a time step of 2 fs.

\subsection{Peptide Sequence Analysis}
\label{method:analysis}
Physicochemical properties like aromaticity, Eisenberg hydrophobicity, charge, charge density, aliphatic index, hydrophobic moment, hydrophobic ratio, isoelectric point, and instability index were estimated using the GlobalAnalysis method in modLAMP \cite{muller2017modlamp}. Protparam tool from Expasy (https://web.expasy.org/protparam) was used to estimate the grand average of hydropathicity (GRAVY) score. 

Pairwise sequence similarity was estimated using a global alignment method, the PAM30 matrix\cite{yu2003compositional}, a gap open penalty of -9, and a gap extension penalty of -1 using Pairwise2 function of Biopython package \cite{cock2009biopython}. Higher positive values indicate better similarity. To check the correspondence between sequence similarity and Euclidean distance in $\rvz$-space, a random set of sequence encodings were first selected, and then sequence similarity and  $\rvz$-distance with their close latent space neighbors were estimated.
Sequence similarity with respect to a sequence database was estimated using ``blastp-short'' command from NCBI BLAST sequence similarity search tool \cite{madden2013blast, altschul1990basic}  was used to query generated short sequences by using a word size of 2,  the PAM30 matrix \cite{yu2003compositional}, a gap open penalty of -9, a gap extension penalty of -1, threshold of 16, $comp\_based\_stats$ set to 0 and window size of 15. Alignment score (Bit score), Expect value (E-value), percentage of alignment coverage, percentage of identity, percentage of positive matches or similarity,  percentage of alignment gap were used for analyzing sequence novelty.  
The E-value is a measure of the probability of the high similarity score occurring by chance when searching a database of a particular size.  E-values decrease exponentially as the score of the match increases.
For the search against patented sequences, we used the PATSEQ database \cite{Patseq}.

\subsection{Wet Lab Experiments}
\label{sec:exptdetails}

\subsubsection{MIC Measurement}
All of the peptides were amidated at their C-terminus to remove the negative charge of the C-terminal carboxyl group. 
Antimicrobial activity of the best AMP hits was evaluated against a broad spectrum of bacteria for minimum inhibitory concentration (MIC), which include Gram-positive \textit{Staphylococcus aureus} (ATCC 29737), Gram-negative \textit{Escherichia coli} (ATCC 25922), \textit{Pseudomonas aeruginosa} (ATCC 9027) and multi-drug resistant \textit{K. pneumoniae} (ATCC 700603), which produce beta-lactamase SHV 18. The broth microdilution method was used to measure MIC values of the AMPs, and the detailed protocol was reported previously \cite{chin2018macromolecular,ng2013synergistic}.
\subsubsection{Hemolytic Activity}
The selectivity of AMPs towards bacteria over mammalian cells was studied using rat red blood cells (rRBCs), which were obtained from the Animal Handling Unit of Biomedical Research Center, Singapore. Negative control: Untreated rRBC suspension in phosphate-buffered saline (PBS); Positive control: rRBC suspension treated with 0.1\% Triton X. Percentage of hemolysis of rRBCs was obtained using the following formula:
\begin{equation}
    Hemolysis (\%)=\frac{O.D._{576 nm} \textit{ of treated samples } - O.D._{576 nm} \textit{ of negative control}}{O.D._{576 nm} \textit{ of positive samples} - O.D._{576 nm} \textit{ of negative control}}
\end{equation}

\subsubsection{Acute in Vivo Toxicity Analysis}
The animal study protocols were approved by the Institutional Animal Care and Use Committee of Biological Resource Center, Agency for Science, technology, and Research (A*STAR), Singapore. LD50 values of the AMPs, the dose required to kill 50\% mice, were determined using a previously reported protocol \cite{liu2017highly}. Specifically, female Balb/c mice (8 weeks old, 18-22 g) were employed. AMPs were dissolved in saline and administered to mice by intraperitoneal (i.p.) injection at various doses. Mortality was monitored for 14 days post-AMP administration, and LD50 was estimated using the maximum likelihood method. 
\subsubsection{CD Spectroscopy} The peptides were dissolved at 0.5 mg/mL in either deionized water or deionized water containing 25 mM SDS surfactant. It forms anioic micelles in aqueous solution, which mimic the bacterial membrane. The CD spectra were measured using a CD spectropolarimeter from Jasco Corp. J-810 at room temperature and a quartz cuvette with 1 mm path length. The spectra were acquired by scanning from 190 to 260 nm at 10nm/min after subtraction with the spectrum of the solvent.

\subsection{Mechanism study using Confocal microscopy}
The mechanism study was performed on an FV3000 Olympus confocal laser scanning microscope. E. coli was grown according to the MIC measurement. Briefly, 500 $\mu$l of $10^8$ CFU/ml of bacteria was prepared and placed in a centrifuged tube and 500 $\mu$l of peptide and polymyxin B at 4xMIC concentration was added. The mixture was incubated at 37 $^{\circ}$C and shaken at 100 rpm for 2 h. The sample was centrifuged at 4000 rpm for 10 min and supernatant was discarded before fresh 1 ml PBS was added. After washing for 3 times, the pallet was re-suspended with 500 $\mu$l of glutaraldehyde (2.5\% v/v) for 30 min with occasional mixing. The sample was again washed for 3 times and re-suspended in 100 $\mu$l of PBS and applied onto a poly-L-lysine coated slide. The bacteria was left to attach for an hour before the unattached bacteria was washed away with PBS. The slides was left to air dry before imaging using the FV3000 confocal microscope.

\subsection{Resistance Study}
Drug resistance in the E. coli can be induced by treating repeatedly with peptides and imipenem at the sub MIC concentration (1/2 MIC). MIC experiment was performed as stated earlier using the micro dilution method. The next generation of E. coli can be initiated by growing the bacteria suspension from the sub MIC concentration and performing another MIC assay.  After 18h inoculation, the bacteria grown from sub MIC concentration of the tested peptides/antibiotics was assayed for MIC. This will continue until we have reached 25 generations. The change in MIC was documented by taking the normalized value of MIC at generation X against the MIC at generation 1.

\begin{table}[ht]
    \centering
    \begin{tabular}{|c|c|c|c|c|c|c|}
        \toprule
        \textbf{E-value} & $<=0.001$ & $<=0.01$ & $<=0.1$ & $<=1$ & $<=10$ & $>10$ \\ \hline   
         Labeled & 9.36 & 3.42 & 9.70 & 29.59 & 29.88 & 18.05  \\
         Unlabeled & 5.16 & 4.05 & 9.07 & 30.97 & 36.65 & 14.08  \\
         \bottomrule
    \end{tabular}
    \caption{Percentage of CLaSS-generated AMP sequences in different categories of Expect value (E-value). E-value for the match with the highest score was considered, as obtained by performing BLAST similarity search against AMP-labeled training sequences (top row) and unlabeled training sequences (bottom row).  
}
    \label{tab:Evalue}
\end{table}{}


\begin{figure}
\centering

\subcaptionbox{Fraction of unique $k$-mer present in different datasets. $k$ = 3-6. The mean and standard errors were estimated on three different sets, each consisting of 3000 randomly chosen samples.}{\begin{tabular}{c|c|c|c}
\toprule
$k$ & Generated AMP & AMP-labeled Training &  Unlabeled Training \\ \hline
3 & $0.299 \pm 0.006$ & $0.223 \pm 0.006$ &  $0.110 \pm 0.005$ \\
4 & $0.771 \pm 0.004$ & $0.607 \pm 0.002$ &  $0.777 \pm 0.001$ \\
5 & $0.947 \pm 0.002$ & $0.706 \pm 0.002$ &  $0.965 \pm 0.000$ \\
6 & $0.978 \pm 0.001$ & $0.742 \pm 0.002$ &  $0.983 \pm 0.001$ \\
\bottomrule
\end{tabular}}%
\centering
\hfill
\subcaptionbox{Top 3 $k$-mers ($k$ = 3 and 4) and their corresponding frequency in generated AMPs,  training AMPs, and unlabeled training sequences. The estimated standard deviation values were close to zero.}{\begin{tabular}{l|ll|ll|ll}
\toprule
\multirow{2}{*}{Kmers} & \multicolumn{2}{|c|}{Generated AMP} & \multicolumn{2}{|c|}{AMP-labeled Training} & \multicolumn{2}{|c}{Unlabeled Training} \\
\cmidrule(lr){2-3} \cmidrule(lr){4-5} \cmidrule(lr){6-7}
                                    & top-3         & Frequency        & top-3         & Frequency         & top-3         &  Frequency       \\ \hline
\multirow{3}{*}{3}                  & KKK       & $0.011$           &  KKK          & $0.006$           & LLL          &  $0.002$          \\
                                    & LKK       & $0.007$           &  KKL          & $0.004$           & LAA          &  $0.001$         \\
                                    & KLK       & $0.007$           &  GLL          & $0.004$           & RRR          &  $0.001$          \\
\cmidrule{1-7}                   
\multirow{3}{*}{4}                  & KKKK       & $0.005$           &  KKKK          & $0.003$           & PLDL          &  $0.001$          \\
                                    & KLKK       & $0.004$           &  KKLL          & $0.002$           & LDLA          &  $0.001$          \\
                                    & KKLK       & $0.003$           &  KLLK          & $0.002$           & NFPL          &  $0.001$          \\
\bottomrule
\end{tabular}}%

\hfill
\subcaptionbox{Physicochemical properties, such as charge, charge density, aliphatic index, aromaticity, hydrophobicity, hydrophobic moment, hydrophobic ratio, instability index, were estimated on unlabeled training, AMP-labeled training, and generated AMP sequences using CLaSS. Mean, and standard deviation were estimated on three different sets, each consisting of 3000 randomly chosen samples. }{\begin{tabular}{l|rrr}
\toprule
{} & Generated AMP & AMP-labeled Training & Unlabeled Training \\ \hline
Charge & $2.695 \pm 0.039$ & $3.074 \pm 0.238$ &  $1.172 \pm 0.218$\\
Charge Density & $0.002 \pm 0.000$ &  $0.002 \pm 0.000$ &  $0.001 \pm 0.000$\\
Aliphatic Index & $107.814 \pm 0.649$ & $101.232 \pm 1.550$ &  $82.088 \pm 8.440$ \\
Aromaticity & $0.095 \pm 0.002$ & $0.102 \pm 0.002$ &  $0.082  \pm 0.003$ \\
Hydrophobicity & $0.048 \pm 0.007$ & $0.068 \pm 0.002$ & $0.052 \pm 0.032$\\
Hydrophobic Moment & $0.331 \pm 0.000$ & $0.335 \pm 0.019$ & $0.222 \pm 0.003$ \\
Hydrophobic Ratio & $0.446 \pm 0.001$ & $0.431 \pm 0.010$ &  $0.425 \pm 0.013$ \\
 \bottomrule
\end{tabular}
}%
\caption{Comparison of CLaSS-generated AMPs with AMP-labeled and unlabeled peptides.}
\label{tab:combined-comp}
\end{figure}



\begin{table}[ht]
\centering

\begin{tabular}{|l|c|c|c|c|}
\hline

\multicolumn{1}{|c|}{\textbf{Sequence}} & \textbf{Positive Residue} & \textbf{Binding time (ns)} & \textbf{Mean}  & \textbf{Variance } \\ \hline
YLRLIRYMAKMI & 3 & 210 & 6.45 & 1.27 \\ \hline
FPLTWLKWWKWKK & 4 & 90 & 5.90 & 1.39 \\ \hline

HILRMRIRQMMT & 3 & 17 & 7.84 & 1.44 \\ \hline
ILLHAILGVRKKL & 3 & 105 & 7.16 & 1.19 \\ \hline
YRAAMLRRQYMMT & 3 & 19 & 8.79 & 1.25 \\ \hline
HIRLMRIRQMMT & 3 & 493 & 8.38 & 1.50 \\ \hline
HIRAMRIRAQMMT & 3 & 39 & 7.20 & 1.39 \\ \hline
KTLAQLSAGVKRWH & 3 & 177 & 7.62 & 1.46 \\ \hline

HILRMRIRQGMMT & 3 & 62 & 8.37 & 1.53 \\ \hline
HRAIMLRIRQMMT & 3 & 297 & 7.46 & 1.35 \\ \hline
EYLIEVRESAKMTQ & 2 & 150 & 6.65 & 1.79 \\ \hline
GLITMLKVGLAKVQ & 2 & 341 & 8.34 & 1.58 \\ \hline
YQLLRIMRINIA & 2 & 239 & 6.29 & 1.71 \\ \hline
VRWIEYWREKWRT & 4 & 125 & 6.41 & 1.28 \\ \hline
LIQVAPLGRLLKRR & 4 & 37 & 6.52 & 1.24 \\ \hline
YQLRLIMKYAI & 2 & 192 & 7.75 & 1.86 \\ \hline
HRALMRIRQCMT & 3 & 80 & 9.15 & 1.27 \\ \hline
GWLPTEKWRKLC & 3 & 227 & 6.11 & 1.63 \\ \hline
YQLRLMRIMSRI & 3 & 349 & 8.28 & 1.80 \\ \hline

LRPAFKVSK & 3 & 151 & 7.73 & 1.85 \\
\hline
\end{tabular}
\caption{Physics-derived features such as mean and variance of the number of contacts between positive amino acids and membrane beads (that are found to be associated with antimicrobial function in this study), as extracted from CGMD simulations of peptide membrane interactions for the top 20
AI-designed AMP sequences.}
\label{tab:20P-sim}
\end{table}

\begin{table}[ht]
\centering
\begin{tabular}{|l|c|c|r|c|}
\toprule
\multicolumn{1}{|c|}{\textbf{Sequence}} & \textbf{Length} & \textbf{Charge} & \textbf{H} & \textbf{$\mu$H} \\ \hline 
YLRLIRYMAKMI & 12 & 3.99 & 0.08 & 0.79  \\ \hline 
FPLTWLKWWKWKK & 13 & 5.00 & 0.05 & 0.20 \\ \hline 
HILRMRIRQMMT & 12 & 4.10 & -0.25 & 0.36 \\ \hline 
ILLHAILGVRKKL & 13 & 4.09 & 0.27 & 0.33 \\ \hline 
YRAAMLRRQYMMT & 13 & 3.99 & -0.28 & 0.06 \\ \hline 
HIRLMRIRQMMT & 12 & 4.10 & -0.25 & 0.16 \\ \hline 
HIRAMRIRAQMMT & 13 & 4.10 & -0.22 & 0.24  \\ \hline 
KTLAQLSAGVKRWH & 14 & 4.09 & -0.08 & 0.49  \\ \hline 

HILRMRIRQGMMT & 13 & 4.10 & -0.19 & 0.27  \\ \hline 
HRAIMLRIRQMMT & 13 & 4.10 & -0.18 & 0.41  \\ \hline 
EYLIEVRESAKMTQ & 14 & 0.00 & -0.16 & 0.26 \\ \hline 
GLITMLKVGLAKVQ & 14 & 3.00 & 0.37 & 0.28 \\ \hline 
YQLLRIMRINIA & 12 & 3.00 & 0.11 & 0.38 \\ \hline 
VRWIEYWREKWRT & 13 & 3.00 & -0.41 & 0.55 \\ \hline 
LIQVAPLGRLLKRR & 14 & 5.00 & -0.12 & 0.38 \\ \hline 
YQLRLIMKYAI & 11 & 2.99 & 0.18 & 0.40 \\ \hline 
HRALMRIRQCMT & 12 & 4.03 & -0.34 & 0.56 \\ \hline 
GWLPTEKWRKLC & 12 & 2.93 & -0.13 & 0.33 \\ \hline 
YQLRLMRIMSRI & 12 & 4.00 & -0.17 & 0.64 \\ \hline 

LRPAFKVSK & 9 & 4.00 & -0.17 & 0.70 \\ 
\bottomrule
\end{tabular}

\caption{Physico-chemical properties for the top 20 AI-designed AMP sequences.}
\label{tab:sicgmd}
\end{table}

\begin{table}[ht]
\centering
\begin{tabular}{|l|c|c|}
\hline

\multicolumn{1}{|c|}{\textbf{Sequence}} & \textbf{S. aureus ($\mu$g/mL)} & \textbf{E.Coli ($\mu$g/mL)} \\ \hline
YLRLIRYMAKMI-CONH2  & 7.8 & 31.25 \\ \hline
FPLTWLKWWKWKK-CONH2  & 15.6 & 31.25 \\ \hline
HILRMRIRQMMT-CONH2  & $>$1000 & $>$1000 \\ \hline
ILLHAILGVRKKL-CONH2  & 250 & 250 \\ \hline
YRAAMLRRQYMMT-CONH2  & $>$1000 & $>$1000 \\ \hline
HIRLMRIRQMMT-CONH2  & $>$1000 & $>$1000 \\ \hline
HIRAMRIRAQMMT-CONH2  & $>$1000 & 1000 \\ \hline
KTLAQLSAGVKRWH-CONH2  & $>$1000 & $>$1000 \\ \hline

HILRMRIRQGMMT-CONH2  & $>$1000 & $>$1000\\ \hline
HRAIMLRIRQMMT-CONH2  & $>$1000 & $>$1000 \\ \hline
EYLIEVRESAKMTQ-CONH2  & $>$1000 & $>$1000 \\ \hline
GLITMLKVGLAKVQ-CONH2  & $>$1000 & $>$1000 \\ \hline
YQLLRIMRINIA-CONH2  & $>$1000 & $>$1000 \\ \hline
VRWIEYWREKWRT-CONH2  & $>$1000 & $>$1000 \\ \hline

YQLRLIMKYAI-CONH2  & 125 & 125 \\ \hline
HRALMRIRQCMT-CONH2  & 1000 & 1000 \\ \hline
GWLPTEKWRKLC-CONH2  & 1000 & $>$1000 \\ \hline
YQLRLMRIMSRI-CONH2  & 250 & 500\\ \hline
FFPLPAISTELKRL-CONH2  & $>$1000 & $>$1000 \\ \hline
LIQVAPLGRLLKRR-CONH2  & $>$1000 & 1000 \\ \hline

LRPAFKVSK-CONH2  & $>$1000 & $>$1000 \\ \hline
\end{tabular}
\caption{Broad-spectrum MIC values of top 20 AI-designed AMP Sequences}
\label{tab:20P_MIC}
\end{table}

\begin{table}[ht]
\centering
\begin{tabular}{|l|c|c|}
\hline
\multicolumn{1}{|c|}{\textbf{Sequence}}& \textbf{S. aureus ($\mu$g/mL)} & \textbf{E.Coli ($\mu$g/mL)} \\ \hline
AMLELARIIGRR-CONH2  & $>$1000 & $>$1000 \\ \hline
IPRPGPFVDPRSR-CONH2  & $>$1000 & $>$1000 \\ \hline
VAKVFRAPKVPICP-CONH2  & $>$1000 & $>$1000 \\ \hline
FPSFTFRLRKWKRG-CONH2 & 62.5 & 62.5 \\ \hline
RPPFGPPFRR-CONH2  & $>$1000& $>$1000 \\ \hline
WEEMDSLRKWRIWS-CONH2  & $>$1000 & $>$1000 \\ \hline
RRQAQEVRGPRH-CONH2  & $>$1000 & $>$1000 \\ \hline
KKKKPLTPDFVFF-CONH2  & $>$1000 & $>$1000 \\ \hline
TRGPPPTFRAFR-CONH2  & $>$1000 & $>$1000 \\ \hline
LALHLEALIAGRR-CONH2  & 250 & $>$1000\\ \hline
\end{tabular}
\caption{Broad-spectrum MIC values of 11 AI-designed Non-AMP Sequences}
\label{tab:11N-MIC}
\end{table}

\renewcommand{\thesubfigure}{\Alph{subfigure}}

\end{document}